\PassOptionsToPackage{table}{xcolor}
\PassOptionsToPackage{numbers,sort&compress}{natbib}
\documentclass[runningheads]{llncs}

\usepackage[preprint]{colm} 

\setcitestyle{authoryear}

\usepackage{amsmath}
\usepackage{amssymb}
\usepackage{graphicx}
\usepackage{booktabs}
\ifdefined\pdfcompresslevel\else\newcount\pdfcompresslevel\fi
\ifdefined\pdfobjcompresslevel\else\newcount\pdfobjcompresslevel\fi
\ifdefined\pdfoptionpdfminorversion\else\newcount\pdfoptionpdfminorversion\fi
\ifdefined\pdfminorversion\else\newcount\pdfminorversion\fi
\ifdefined\pdfgentounicode\else\newcount\pdfgentounicode\fi
\ifdefined\pdfglyphtounicode\else\def\pdfglyphtounicode#1#2{}\fi
\usepackage[accsupp]{axessibility}
\usepackage{xltabular} 
\usepackage{microtype}
\usepackage{tabularx}
\usepackage{paralist}
\usepackage{todonotes}
\usepackage{subcaption}
\usepackage{latexsym}
\usepackage{float}
\usepackage{footnote}
\usepackage{enumitem}
\usepackage{bm}
\usepackage{arydshln}
\usepackage{multicol}
\usepackage{multirow}
\usepackage{colortbl}
\usepackage{bbding}
\usepackage{makecell}
\usepackage{mathtools}
\usepackage{imakeidx}
\usepackage{longtable}
\usepackage{tabularx}
\usepackage{wrapfig}
\usepackage{rotating}
\usepackage[edges]{forest}
\usepackage[normalem]{ulem}
\usepackage{CJKutf8}
\ifdefined\XeTeXversion \else \usepackage{awesomebox} \fi
\usepackage[most]{tcolorbox}

\RequirePackage{xspace}
\makeatletter
\DeclareRobustCommand\onedot{\futurelet\@let@token\@onedot}
\def\@onedot{\ifx\@let@token.\else.\null\fi\xspace}

\makeatother

\usepackage{pifont}
\usepackage{lscape}
\usepackage{algorithm}
\usepackage{algpseudocode}

\usepackage{fancyhdr}
\renewcommand{\headrulewidth}{1pt}
\usepackage{fancyvrb}
\usepackage{fvextra}
\usepackage{amsfonts}
\usepackage{wrapfig}

\ifdefined\XeTeXversion
  \usepackage{fontspec}
\fi

\usepackage{hyperref}
\usepackage{orcidlink}

\definecolor{mydarkblue}{rgb}{0,0.08,0.45}
\definecolor{wkblue}{rgb}{0.2, 0.3, 0.6}
\definecolor{meta-color}{rgb}{0.5, 0.5, 0.5}
\definecolor{darkblue}{rgb}{0, 0, 0.5}
\definecolor{geovistagray}{gray}{0.95}
\definecolor{myblue}{rgb}{0.9, 0.1, 0.94}
\definecolor{mygreen}{rgb}{0.64, 0.56, 0.88}
\definecolor{myyellow}{rgb}{0.68, 0.6, 0.1}
\definecolor{fancygreen}{rgb}{0.33, 0.68, 0.20}
\definecolor{salmon}{rgb}{0.94, 0.52, 0.49}
\definecolor{tablegreen}{rgb}{0.82, 0.94, 0.75}
\definecolor{tableblue}{rgb}{0.81, 0.90, 0.94}
\definecolor{tablered}{rgb}{0.97, 0.85, 0.85}
\definecolor{tableorange}{rgb}{0.96, 0.85, 0.81}
\definecolor{bestcolor}{RGB}{210, 222, 239}
\definecolor{secondcolor}{RGB}{234, 239, 247}
\definecolor{thirdcolor}{RGB}{193, 214, 229}
\definecolor{line-blue}{RGB}{243, 248, 252}
\definecolor{line-green}{RGB}{200,242,200}
\definecolor{line-red}{RGB}{255,215,215}
\definecolor{line-gray}{RGB}{242, 242, 242}
\definecolor{sensepurple}{HTML}{5D2DD6}

\newenvironment{itemize*}%
 {\leftmargini=10pt\begin{itemize}%
  \setlength{\itemsep}{0pt}%
  \setlength{\parskip}{0pt}%
  }%
 {\end{itemize}}
\newenvironment{enumerate*}%
 {\begin{enumerate}%
  \setlength{\itemsep}{0pt}%
  \setlength{\parskip}{0pt}}%
 {\end{enumerate}}

\setcitestyle{numbers,square,comma}

\usepackage{tikz}
\usetikzlibrary{arrows.meta, positioning, fit, backgrounds}
\definecolor{indigo}{RGB}{75,0,130}

\begin{document}

\title{%
  \begin{center}
    {\LARGE\bfseries Hunyuan3D-Buffalo 1.0}\\[0.4em]
    {\large A Unified Multimodal Model for Scalable 3D Generation, \\ Understanding, and Editing}
  \end{center}%
  \vspace{-0.35cm}
}

\author{%
\makebox[\textwidth][c]{\normalsize\bfseries
Tencent Hunyuan\textsuperscript{*}}}



\maketitle

\begingroup
\renewcommand{\thefootnote}{\fnsymbol{footnote}}
\endgroup

\addtocounter{footnote}{0}

\begin{abstract}
Recent advances in image generation have demonstrated the potential of unified multimodal models that integrate understanding, generation, and editing. However, unified 3D modeling remains constrained by scarce multimodal data, particularly the lack of large-scale and geometrically consistent editing data. To address this limitation, we propose \textbf{Hunyuan3D-Buffalo 1.0}, a unified framework supporting 3D understanding, text-to-3D generation, instruction-guided 3D editing, and text-grounded part generation within a single architecture. To enable scalable training, we construct an 87M-scale 3D multimodal corpus, comprising 25M understanding samples, 50M text-to-3D pairs, and 12M editing pairs generated using Nano3D-v2. Architecturally, the framework combines Hunyuan3D-VLM for semantic, structural, and spatial understanding with Hunyuan3D DiT for high-fidelity 3D synthesis. The VLM provides multimodal semantic conditions for generation, while editing and part generation additionally condition the diffusion process on the source object representation to preserve its overall structure and unedited regions. Extensive experiments show that Hunyuan3D-Buffalo 1.0 achieves state-of-the-art or leading performance on text-to-3D generation and 3D editing benchmarks, while exhibiting strong understanding and part-generation capabilities. Our analysis further shows that both generation and understanding improve editing, demonstrating the effectiveness of unified 3D multimodal training.

\end{abstract}

\vspace{-0.5cm}
\begin{center}
    \begin{figure}[h!]
      \centering
      \includegraphics[width=\linewidth]{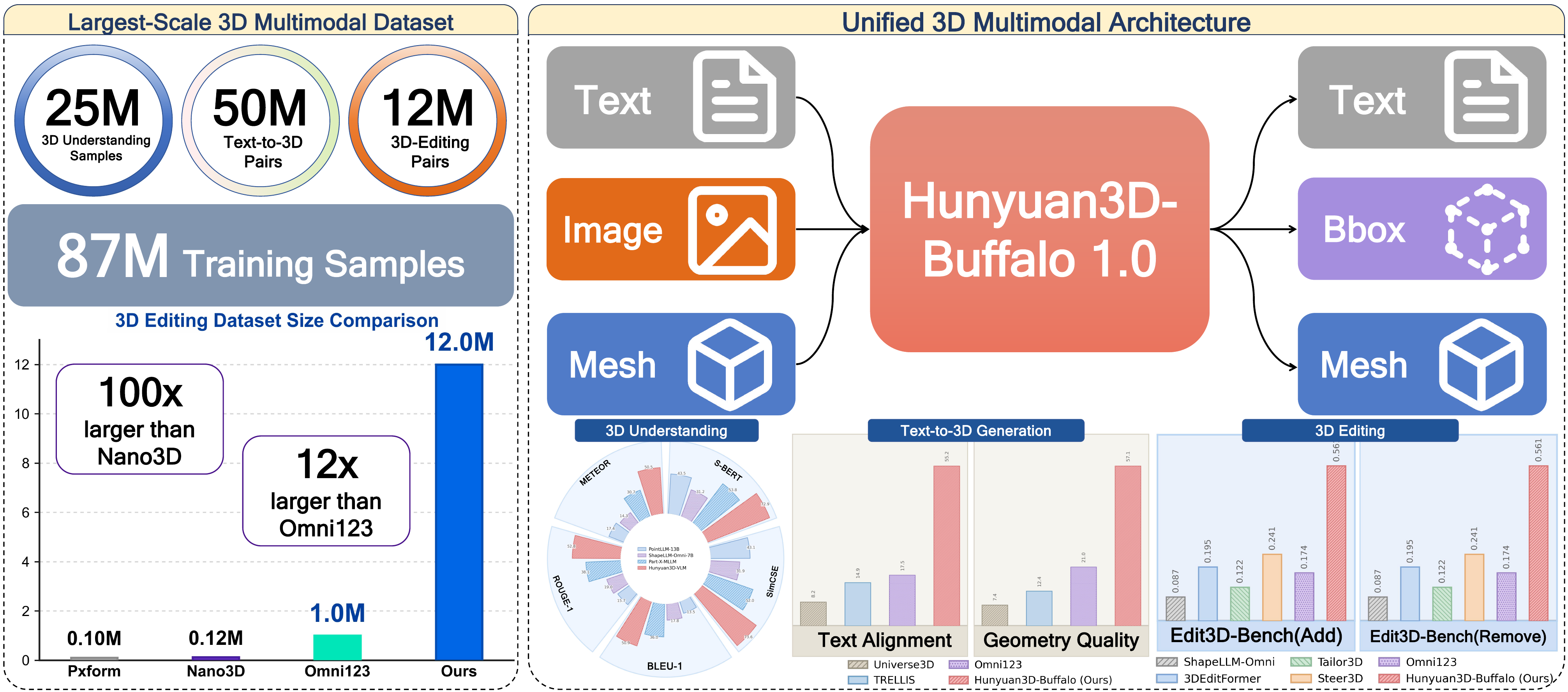}
      \label{fig:Place}
    \end{figure}
\end{center}

\begin{center}

    \begin{figure}[h!]
      \centering
        \includegraphics[width=\linewidth,height=\textheight,keepaspectratio]{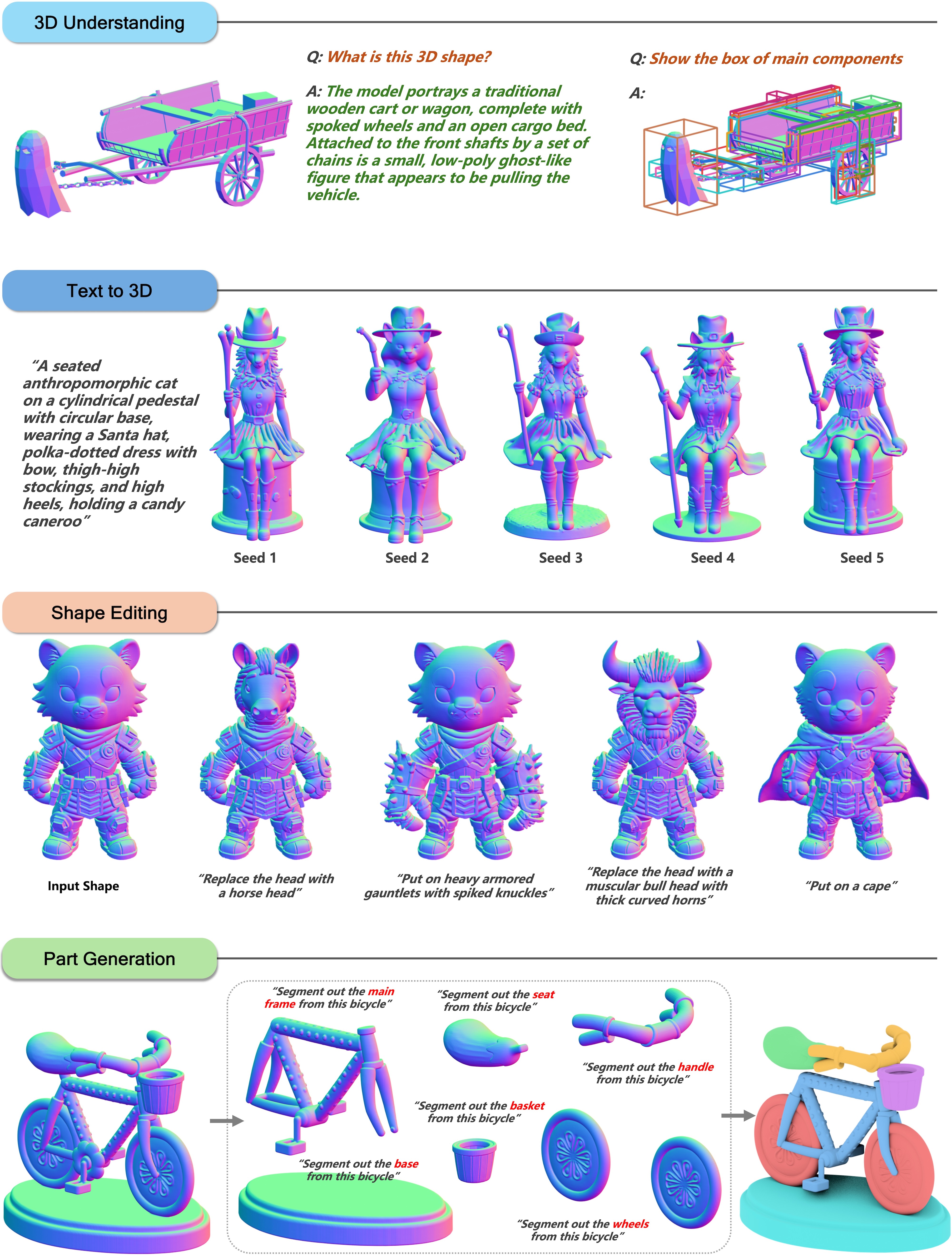}
      \vspace{0.3cm}
      \caption{   \textbf{Hunyuan3D-Buffalo 1.0} is a unified 3D multimodal framework that combines autoregressive modeling with diffusion-based generation, enabling 3D understanding, text-to-3D generation, 3D editing, and text-grounded part generation within a single architecture.}
 
    \end{figure}

\end{center}
\vspace*{\fill}

\newpage
\setcounter{tocdepth}{3}
\setcounter{secnumdepth}{3}
\tableofcontents

 

\newpage

\section{Introduction}

Recent advances in generative artificial intelligence have driven 2D vision models toward unified multimodal systems that integrate understanding, generation, and instruction-guided editing. Models such as GPT-4o, FLUX~\citep{labs2025flux1kontextflowmatching}, Seedream~\citep{gao2025seedream, seedream2025seedream}, Qwen-Image~\citep{wu2025qwen}, Nano-Banana, and the GPT-Image series have shown that visual generation is no longer an isolated synthesis task, but can serve as a general interface for multimodal reasoning, content creation, and interactive editing~\citep{deng2025bagel, chen2025janus, wang2024emu3, wu2025omnigen2, chen2025blip3, cui2025emu35nativemultimodalmodels}. More recently, works such as Vision-Banana further suggest that dense perception and part-level understanding can also be formulated through generative modeling, revealing the potential of unified architectures to transfer capabilities across tasks.

However, analogous progress in the 3D domain remains limited. Unlike 2D images, 3D assets are much harder to collect, annotate, and edit at scale. In particular, large-scale and geometrically consistent 3D editing data is scarce, making it difficult to train models that can modify existing 3D assets while preserving identity, structure, and unedited regions. As a result, 3D understanding models~\citep{xu2024pointllm, qi2024shapellm, qi2024gpt4point}, 3D generation models~\citep{zhang2024clay, xiang2025structured, li2025triposg, lai2025hunyuan3d}, and 3D editing models~\citep{li2025voxhammer, ye2025nano3d, chen2023shapeditorinstructionguidedlatent3d} are still largely developed as separate systems. This fragmentation prevents the model from learning a unified semantic--visual--geometric representation, and leaves it unclear whether 3D understanding, generation, and editing can mutually reinforce one another under a unified training paradigm.

To address the data bottleneck, we first construct a large-scale 3D multimodal training corpus with 87M samples, covering 25M 3D understanding samples, 50M text-to-3D pairs, and 12M 3D editing pairs. A key component of this data engine is Nano3D-v2, an agent-based 3D editing data construction algorithm that extends Nano3D~\citep{ye2025nano3d}. Nano3D-v2 combines anchor-view selection, learned 3D edit-region localization, voxel-level editing, fine-grained geometry and texture refinement, and VLM-based annotation and filtering. This pipeline enables scalable construction of high-quality, geometrically consistent editing pairs, substantially mitigating the scarcity of 3D editing supervision.

On top of this data foundation, we propose Hunyuan3D-VLM, a 3D vision-language model designed for fine-grained semantic, structural, and spatial understanding of 3D objects. Hunyuan3D-VLM encodes both geometric structure and appearance cues, enabling object-level captioning, part-level question answering, 3D grounding, edit-instruction synthesis, and edit-outcome reasoning. By equipping the model with explicit 3D perception and grounding ability, Hunyuan3D-VLM provides the semantic and spatial reasoning foundation required for unified 3D multimodal modeling.

Building upon Hunyuan3D-VLM, we further introduce \textbf{Hunyuan3D-Buffalo 1.0}, a unified 3D multimodal framework that combines autoregressive modeling with diffusion-based 3D generation. The framework connects Hunyuan3D-VLM with a generative 3D-DiT backbone initialized from Hunyuan3D-2.1~\citep{hunyuan3d2025hunyuan3d}, allowing high-level multimodal reasoning to guide 3D synthesis and editing through a unified conditional interface. As a result, Hunyuan3D-Buffalo 1.0 supports 3D understanding, text-to-3D generation, instruction-guided 3D editing, and text-grounded part generation within a single architecture. Inspired by the generative formulation of dense perception in Vision-Banana, we also cast part generation as a native instruction-following and conditional generation task, enabling the model to decompose and generate language-referred 3D parts without relying on a specialized part-generation pipeline.

Extensive experiments demonstrate that Hunyuan3D-Buffalo 1.0 achieves state-of-the-art or leading performance on both text-to-3D generation and 3D editing benchmarks, while also showing strong 3D understanding and part-generation abilities. More importantly, our analysis reveals two clear cross-task synergies. First, stronger text-to-3D generation improves 3D editing, as a better generative prior leads to more complete and natural edited geometry. Second, stronger 3D understanding improves 3D editing, as fine-grained semantic and spatial reasoning helps the model localize editing regions, interpret instructions, and preserve unedited areas. These findings show that unified 3D multimodal training is not merely a combination of multiple tasks, but can induce meaningful capability transfer across understanding, generation, and editing.

In summary, our contributions are as follows:

1. We construct a large-scale 3D multimodal training corpus with 87M samples, including 25M 3D understanding samples, 50M text-to-3D pairs, and 12M 3D editing pairs. To build the editing data, we introduce \textbf{Nano3D-v2}, an agent-based 3D editing data construction algorithm that produces high-quality and geometrically consistent editing pairs at scale.

2. We propose Hunyuan3D-VLM, a 3D vision-language understanding architecture capable of fine-grained semantic, structural, and spatial understanding of 3D objects, providing a strong foundation for 3D grounding, part-level reasoning, and edit-aware understanding.

3. We develop \textbf{Hunyuan3D-Buffalo 1.0}, a unified 3D multimodal framework that combines autoregressive modeling with diffusion-based generation, enabling 3D understanding, text-to-3D generation, 3D editing, and text-grounded part generation within a single architecture.

4. Hunyuan3D-Buffalo 1.0 achieves state-of-the-art or leading performance on text-to-3D generation and 3D editing benchmarks. Through unified training, we further reveal two synergistic relationships among 3D multimodal tasks: generation improves editing, and understanding improves editing.

\section{Related Work}
\subsection{3D Generation}
Recent advances in 3D content generation have explored three major paradigms.
Early works,
represented by DreamFusion~\cite{poole2022dreamfusion},
mainly rely on optimization-based methods that distill visual priors from pretrained 2D diffusion models into 3D representations,
enabling text-to-3D generation without large-scale 3D supervision~\cite{lin2023magic3d,chen2023fantasia3d,shi2023mvdream,qiu2024richdreamer,wu2024consistent3d,wang2023prolificdreamer,tang2023dreamgaussian,yi2024gaussiandreamer,ye2024dreamreward,liu2025dreamreward}.
To improve generation efficiency and multi-view consistency,
subsequent works,
represented by 3DShape2VecSet~\cite{zhang20233dshape2vecset} and TRELLIS~\cite{xiang2025structured},
construct 3D-native VAE and train generative models in 3D latent spaces,
enabling feed-forward generation with faster inference and improved geometric fidelity~\cite{zhao2023michelangelo,li2025triposg,he2025sparseflex,li2025sparc3d,li2024craftsman3d,hunyuan3d2025hunyuan3d,lai2025lattice,chen2025ultra3d}.
Another line of work,
represented by MeshAnything~\cite{chen2024meshanything},
focuses on low-poly mesh generation by tokenizing vertices and faces and modeling them with autoregressive or flow-based generative models,
aiming to produce compact and production-friendly 3D meshes
~\cite{chen2025meshanything,weng2025scaling,tang2024edgerunner,weng2024pivotmesh,hao2024meshtron,kim2025fastmesh,zhao2025deepmesh,wang2026polyflow,li2026meshflow,zhao2026lato,wang2026face}.
Despite these advances,
achieving high fidelity,
structural consistency,
and controllable editing remains an open challenge.

\subsection{3D Editing}

Text-guided 3D editing aims to modify existing 3D assets according to natural language instructions while preserving unedited regions and maintaining 3D consistency. Early methods typically rely on per-instance optimization, using Score Distillation Sampling (SDS) or pretrained 2D diffusion priors to align 3D representations with editing instructions~\cite{haque2023instruct,chen2023shapeditorinstructionguidedlatent3d,sella2023voxetextguidedvoxelediting,zhuang2024tipeditoraccurate3deditor,liu2024makeyour3dfastconsistentsubjectdriven,dong2024interactive3dcreatewantinteractive,zhang2025scenelanguagerepresentingscenes,dinh2025geometrystyle3dstylization}. Another line of work edits rendered 2D views and subsequently fuses or reconstructs them into edited 3D assets~\cite{mvedit,qi2024tailor3dcustomized3dassets,cao2024mvinpainterlearningmultiviewconsistent,erkoç2024preditor3dfastprecise3d,gao20243dmesheditingusing,barda2024instant3ditmultiviewinpaintingfast,li2025cmdcontrollablemultiviewdiffusion,zheng2025pro3deditorprogressiveviewsperspective,baron2025editp233deditingpropagation}. While the former incurs costly per-instance optimization, the latter depends heavily on projection and reconstruction consistency and may introduce cross-view inconsistencies or geometric distortions. More recently, training-free methods, exemplified by Nano3D~\cite{ye2025nano3d} and VoxHammer~\cite{li2025voxhammer}, directly edit the latent or structured representation spaces of pretrained 3D generative models by reusing frozen 3D priors through inversion, latent replacement, flow-based editing, or agentic tool chains~\cite{zhou2025anchorflow,chi2026vinedresser3d,sella2026proxe,lim2026tango,hsiao2026vecsetedit,liu2026velocity,cai2025native}. These methods demonstrate the editability of pretrained 3D priors, but their performance can remain unstable across diverse objects, edit types, and instructions. In parallel, training-based approaches construct paired or self-generated 3D editing data and lightly fine-tune pretrained 3D generators to learn feed-forward edit transformations~\cite{xia2025editformer,gat2026shapeup,hu2026easy3e,chen2026omni3dedit,xu2026beyond,weng2026partflow,ma2025steer3d,yin2026editverse3d}. Despite these advances, achieving precise and efficient 3D editing while ensuring structural consistency and faithful preservation of unedited regions remains challenging.

\subsection{3D Part Generation}
3D part generation aims to produce structured assets where each semantic component is a distinct mesh.
Early methods~\cite{mo2019partnet,yan2024frankenstein} are restricted to fixed part taxonomies,
while recent multi-stage pipelines depend on 2D segmentation and suffer from view-dependent inconsistencies.
More recent 3D-native approaches adopt multi-diffusion-path architectures that synthesize parts through coordinated diffusion branches~\cite{yang2025omnipart,yan2025xpart,ma2025p3,li2025moca,zhu2026cubepart,hunyuan3dstudio2025,li2026seamgpt},
or learn continuous feature fields for part segmentation~\cite{liu2025partfield}.
Another line incorporates physics simulation to ensure inter-part plausibility~\cite{yan2024phycage,yang2026physforge,luo2025bag}.
Despite this progress,
existing methods either assume a fixed part vocabulary or infer part structure implicitly.
CubePart~\cite{zhu2026cubepart} introduces an open-vocabulary,
text-grounded framework for part generation.
In this work,
we demonstrate for the first time that text-grounded part generation can be accomplished within a native 3D multimodal large model,
unifying part generation with other 3D tasks under a single architecture.

\subsection{Unified Multimodal Models}
Recent studies on unified image understanding and generation have developed along several architectural directions.
One line of work,
represented by Chameleon~\cite{Chameleon_Team_Chameleon_Mixed-Modal_Early-Fusion_2024,qu2025tokenflow,wang2024emu3,wu2024vila},
converts images into discrete visual tokens with VQVAE~\cite{van2017neural},
allowing text and images to be modeled within a single autoregressive Transformer under the next-token prediction objective.
Another line adopts a more decoupled design,
as exemplified by MetaQuery~\cite{tong2025metamorph,chen2025blip3,chen2025blip3o,pan2025transfer,wu2025omnigen2,wu2025qwen,zhao2026gem},
where a Multimodal Large Language Model~\cite{bai2025qwen2,touvron2023llama} serves as a semantic encoder for complex inputs and provides conditional features to a Diffusion Transformer for high-fidelity image synthesis.
A third direction seeks tighter integration between understanding and generation.
For example,
Bagel~\cite{deng2025bagel,chen2025janus,cao2025hunyuanimage,huang2025mingunivision} incorporates language modeling and flow-based generation within a unified Transformer backbone,
enabling both tasks to be handled in a more native architecture.
In contrast,
unified modeling in the 3D domain remains relatively underexplored.
Existing efforts such as ShapeLLM-Omni~\cite{ye2025shapellm,bhat2025cube,huang2026cg,yang2026eva01} attempt to incorporate 3D data into a unified framework through VQVAE-based tokenization,
but are still limited in capturing fine-grained geometry and complex spatial structures.


\section{Data curation}
\subsection{Overview}
A central obstacle to unified 3D multimodal modeling is the scarcity of
large-scale, high-quality, and geometrically consistent training data
that jointly covers understanding, generation, and editing. To support
the three capabilities of our framework within a single training
pipeline, we construct a comprehensive 3D data engine that produces three
complementary corpora, summarized in Table~\ref{tab:data_summary}.

First, a \textbf{3D understanding} corpus pairs object point clouds with
instruction-style dialogues spanning captioning, question answering,
grounding, and editing-related tasks, interleaved with general text and
image--text data to preserve the model's language and 2D multimodal
abilities. Second, a \textbf{text-to-3D} corpus provides large-scale
text--asset pairs, built by a fully automated pipeline that synthesizes
compositional prompts, generates and renders the corresponding 3D assets,
and attaches multi-tier, geometry-grounded captions under strict quality
filtering. Third, a \textbf{3D editing} corpus supplies geometrically
consistent (source, edited asset, instruction) triplets, produced by our
\textbf{Nano3D-v2} pipeline and refined through a vision-language-model-based
annotation and filtering procedure. In total, the engine yields
roughly $25$M understanding samples, $50$M text-to-3D pairs, and $12$M
editing pairs. The remainder of this section describes the construction
of each corpus in turn.

\begin{table}[ht]
\centering
\small
\begin{tabular}{@{}llr@{}}
\toprule
Capability & Subset & \#Samples \\
\midrule
3D understanding & Text / image / 3D instruction data & $\sim$25M \\
\midrule
Text-to-3D       & Text--asset pairs                  & $\sim$50M \\
\midrule
\multirow{3}{*}{3D editing}
                 & Human edits                        & $\sim$7M \\
                 & Object edits                       & $\sim$3M \\
                 & Part generation        & $\sim$2M \\
                 & \textit{Subtotal}                  & $\sim$12M\\
\bottomrule
\end{tabular}
\caption{Overview of the full training corpus across 3D understanding,
text-to-3D generation, and instruction-guided 3D editing.}
\label{tab:data_summary}
\end{table} 

\subsection{3D Understanding Data}
\label{sec:data:understanding}

Hunyuan3D-VLM is trained on a large-scale instruction-tuning corpus that
interleaves three modalities: pure text, image--text, and 3D point
cloud--text data. The general text-only and image--text data are drawn
from a mixture of public and in-house instruction corpora and serve to
preserve the model's language reasoning and 2D multimodal perception
during 3D adaptation.
The text-only and image--text splits
contribute approximately $7$M and $3$M conversation samples,
respectively. The 3D point cloud--text data forms the core of the corpus
and accounts for the largest share at roughly $15$M samples, bringing
the total mixture to about $25$M samples.

The 3D split is built upon the part-level and object-level 3D dialogue
data provided by Part-X-MLLM~\cite{wang2025part} and ShapeLLM-Omni~\cite{ye2025shapellm}, together
with data we collect and synthesize in-house from our own asset pool. Across these
sources, the 3D understanding tasks cover:

\begin{itemize}
  \item \textbf{3D captioning}: multi-tier natural-language descriptions
    of an object's geometry, structure, and salient parts.
  \item \textbf{3D question answering}: free-form questions about an
    object's category, attributes, parts, and spatial relations.
  \item \textbf{3D grounding}: localizing referred parts or regions in an
    object, with answers expressed as quantized axis-aligned bounding
    boxes via \texttt{<boxs>}/\texttt{<boxe>} token sequences.
  \item \textbf{Edit-instruction synthesis}: given the source point cloud
    and an original, often coarse editing request, producing a precise,
    executable editing instruction grounded in the object's geometry.
  \item \textbf{Edit-outcome captioning}: given the source object and an
    editing operation, describing the object that results from the edit,
    so that the model learns the correspondence between an editing
    operation and its geometric outcome.
\end{itemize}

\begin{figure}[ht!]
  \centering
  \includegraphics[width=\linewidth]{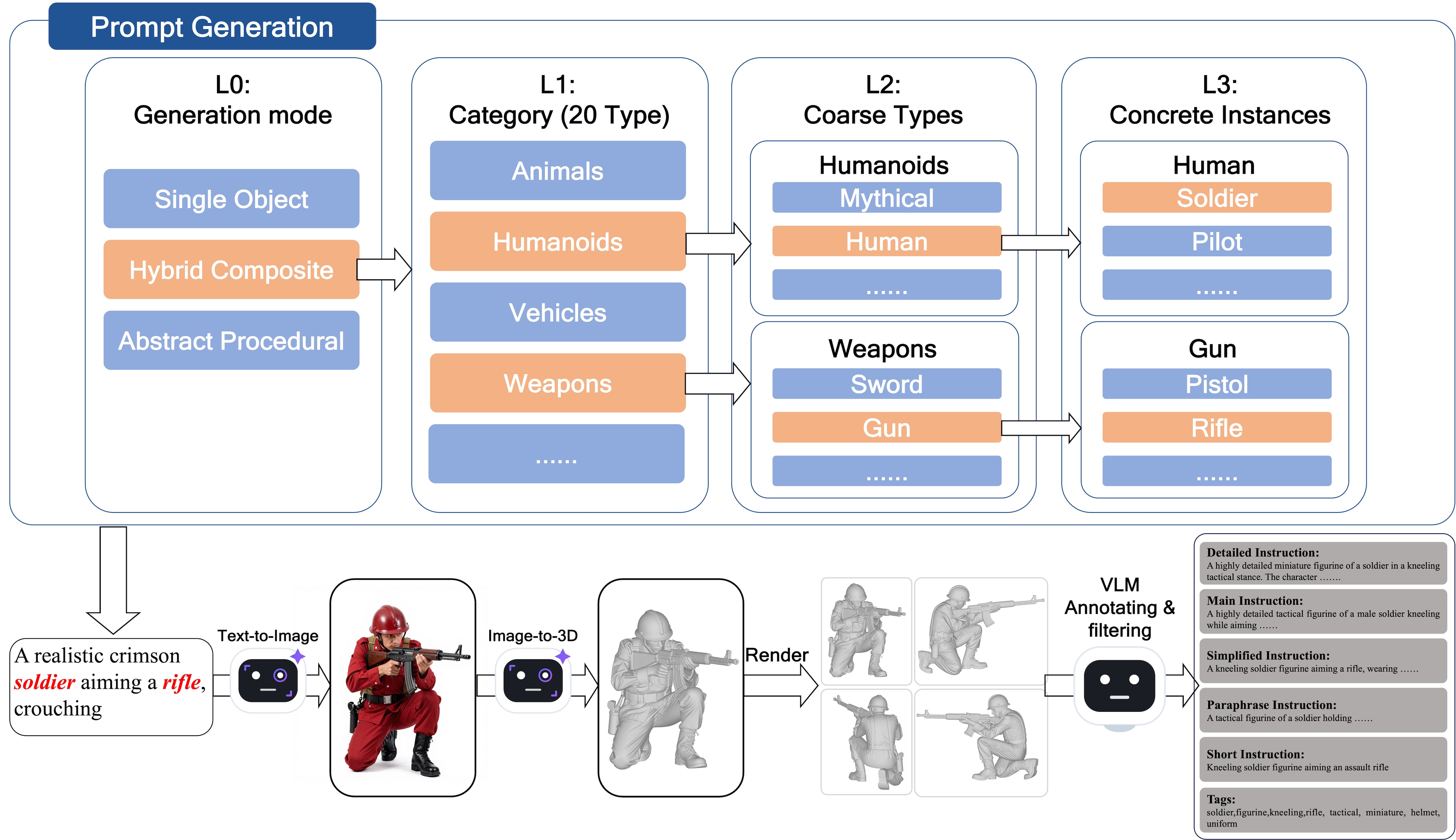}
  \caption{Pipeline of constructing text-to-3D training corpus.}
  \label{fig:t23d-pipeline}
\end{figure}

\subsection{Text-to-3D Data}
We build our text-to-3D training corpus with a fully automated,
five-stage pipeline: (i) hierarchical prompt taxonomy construction;
(ii) compositional prompt synthesis; (iii)  image-to-3D asset generation and
multi-view rendering; (iv) multi-tier captioning and geometry-quality
scoring; and (v) quality filtering and dataset packaging.
Figure~\ref{fig:t23d-pipeline} gives an overview.

\paragraph{Stage 1: Hierarchical Prompt Taxonomy.}

Prompts are sampled from a four-level taxonomy that decouples
\emph{what} to generate from \emph{how} it is composed:

\begin{itemize}
  \item \textbf{L0 -- Generation mode}: \texttt{object\_single},
    \texttt{hybrid\_composite} (a primary subject interacting with a
    secondary object), and \texttt{abstract\_procedural}. Mode weights
    are category-dependent; e.g.\ \emph{Humanoids} and \emph{Animals}
    are biased toward interaction scenes.
  \item \textbf{L1 -- Category} ($20$ categories, weighted): e.g.\
    \emph{Humanoids}, \emph{Animals}, \emph{Vehicles},
    \emph{Architecture}, \emph{Furniture}, \emph{Weapons},
    \emph{Plants}, \emph{Food}, etc. Weights skew the distribution
    toward the categories that matter most for our target use cases.
  \item \textbf{L2 / L3 -- Subtype and fine-grained type}: each L1 is
    expanded into coarse types (L2) and concrete instances (L3),
    yielding thousands of leaf concepts (e.g.\ \emph{Humanoids
    $\rightarrow$ fantasy warrior $\rightarrow$ samurai}).
\end{itemize}

Each leaf is further decorated with \emph{attributes} sampled from
controlled vocabularies: \emph{style} (15 options, weighted toward
render-friendly looks such as \emph{realistic}, \emph{product},
\emph{3D~cartoon}), \emph{color}, \emph{material}, \emph{condition},
and \emph{pose}. Styles that are unsuitable as image-to-3D inputs
(wireframe, blueprint, watercolor, isometric, etc.) are explicitly
removed, and a per-category override table further constrains which
attributes apply to which concepts.

\paragraph{Stage 2: Compositional Prompt Synthesis.}

A single prompt is assembled by sampling
$(\text{L0}, \text{L1}, \text{L2}, \text{L3})$ and attributes, then
composing a natural, user-style request. For \texttt{hybrid\_composite}
prompts, a \emph{semantic-role}–aware interaction sampler selects a
secondary object that the primary subject can \emph{plausibly} interact
with (e.g.\ a knight \emph{wielding} a sword), governed by role
interaction tables, whitelists, and anthropomorphism guards (e.g.\
real animals are prevented from performing physically implausible,
finger- or speech-requiring actions, and are down-styled away from
photoreal looks when forced into stylized interactions). When no valid
contact template exists, the prompt gracefully degrades to a single
object.

To inflate the effective unique-prompt space without altering
semantics, the synthesizer (a) randomly omits optional attribute axes
(material/pose/condition, each with $\sim\!30\%$ drop probability) and
(b) appends a \emph{quality suffix} drawn component-wise from five
paraphrase axes (view angle, framing, background, subject visibility,
detail level). Each record also carries a shared
\texttt{negative\_prompt}. The resulting per-record schema is:

\begin{verbatim}
{ "L0", "L1", "L2", "L3", "attributes","prompt", "negative_prompt", "secondary"? }
\end{verbatim}

\paragraph{Stage 3: Asset Generation and Rendering.}
\label{sec:data:assets}

Prompts are turned into images and then into 3D assets via an
image-to-3D model, after which each asset is rendered to multiple
canonical views (front-facing, white background) and stored alongside a
sampled surface point cloud. For the edit-augmented split, we run an
automated quality-control step: a \emph{difference-detection} module
compares the pre-edit (background-removed, white-matted) and post-edit
images to produce a binary diff mask, and a \emph{consistency check}
verifies that (a) regions \emph{outside} the mask remain unchanged
between pre- and post-edit (sharpness-aligned to avoid false positives)
and (b) the mask does not cover an implausibly large fraction of the
foreground. Both steps are sharded across machines for million-scale
throughput, and failing samples are discarded.

\paragraph{Stage 4: Multi-Tier Captioning and Geometry-Quality Scoring.}

To attach training captions to each generated asset, we query a
vision-language model (Gemini, via an internal endpoint) with
\emph{the reference image as context} and \emph{the multi-view renders
of the actual generated mesh as the source of truth}. The model is
instructed to (i) describe the rendered untextured white mesh in a
natural text-to-3D request voice---never mentioning color, material,
lighting, or the rendering/generation process, and never comparing
against the reference image---and (ii) assign a strict integer
\emph{geometry-quality} score. Concretely it emits six caption tiers
that are \emph{derived} from the most detailed tier (shorter tiers may
drop information but never add it), plus one score:

\begin{table}[t]
\centering
\begin{tabular}{@{}lll@{}}
\toprule
Tier & Length & Emphasis \\
\midrule
Detailed   & 4--6 sentences ($\le$120 w) & subject, parts, pose, features \\
Main       & 24--30 tokens & structure + key parts \\
Simplified & 15--20 tokens & main parts and fit \\
Paraphrase & 15--20 tokens & reworded Simplified \\
Short      & 6--10 tokens  & subject + $\le$1 attribute \\
Tags       & $\le$8 keywords & disentangled keywords \\
\bottomrule
\end{tabular}
\caption{Six caption tiers produced per asset.}
\label{tab:tiers}
\end{table}
The \texttt{geometry\_quality} score lies in $[0,10]$ and penalizes
truncation/cropping, duplicated parts, detached/floating chunks, broken
topology, implausible proportions, and missing parts, with calibrated
anchors ($0$: severely broken/unusable; $3$: recognizable but multiply
defective or truncated; $7$: mostly clean and fully framed; $10$:
clean, complete, and correctly assembled in every view). The model is
explicitly told to use the full range rather than default to high
scores.

\paragraph{Stage 5: Quality Filtering and Dataset Packaging.}
\label{sec:data:package}

Finally, we parse the model responses, extract the six caption tiers
and the geometry-quality score, and filter assets by a score threshold
(we keep the highest-quality tier, \texttt{geometry\_quality}~$\ge$~%
\,$\tau$, with $\tau{=}10$ for our cleanest split). 
Surviving assets are packaged into the training format as tuples of
(surface point-cloud path, caption dictionary, L1 category), yielding
the final dataset of $\sim$\,$5\!\times\!10^{7}$ high-quality
text--asset pairs.

\subsection{Editing Data}

\begin{figure}[ht!]
  \centering
  \includegraphics[width=\linewidth]{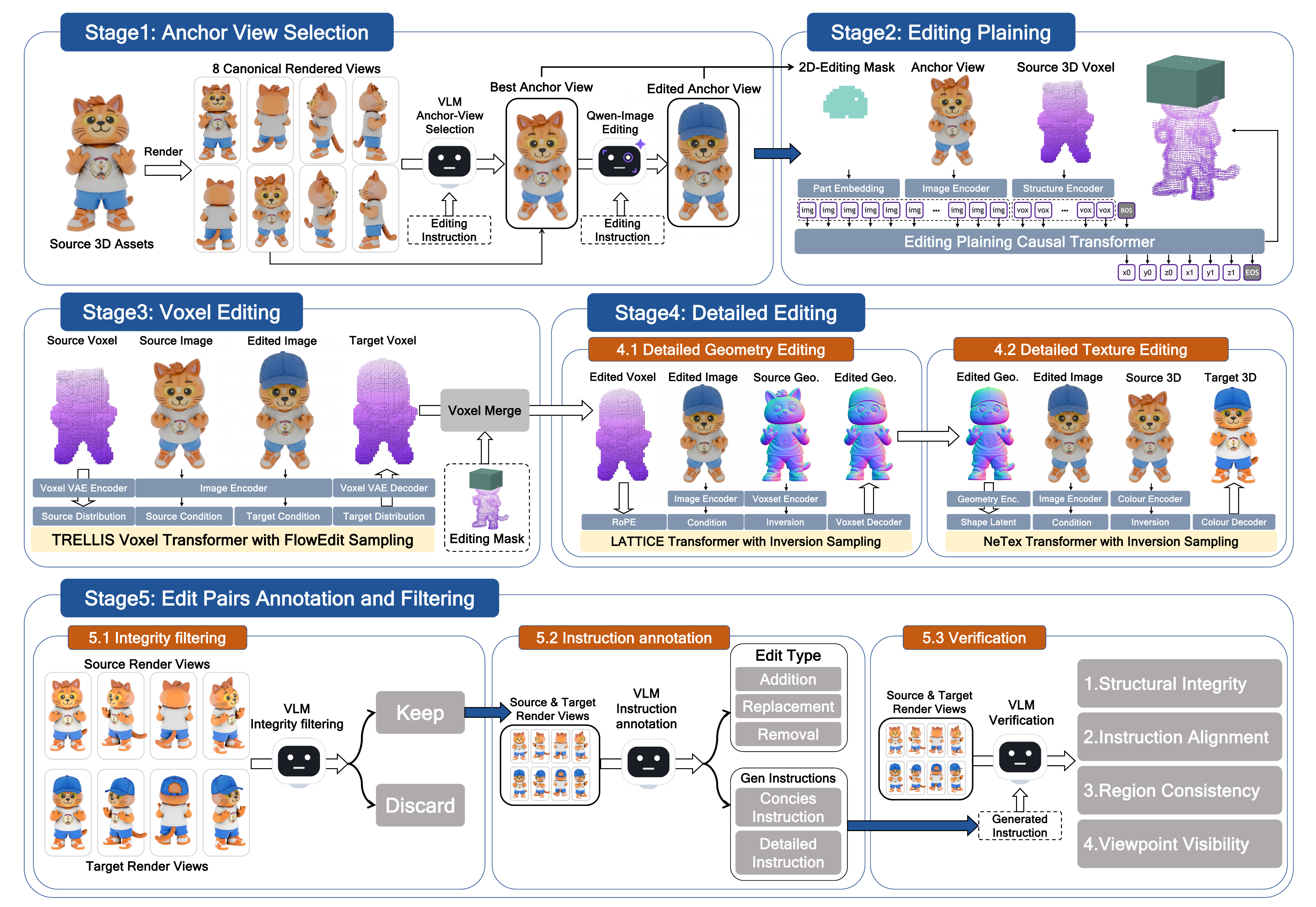}
  \caption{Pipeline of constructing 3D editing training corpus (Nano3D-v2).}
  \label{fig:nano3d-v2-pipeline}
\end{figure}

\begin{figure}[ht!]
  \centering
  \includegraphics[width=\linewidth]{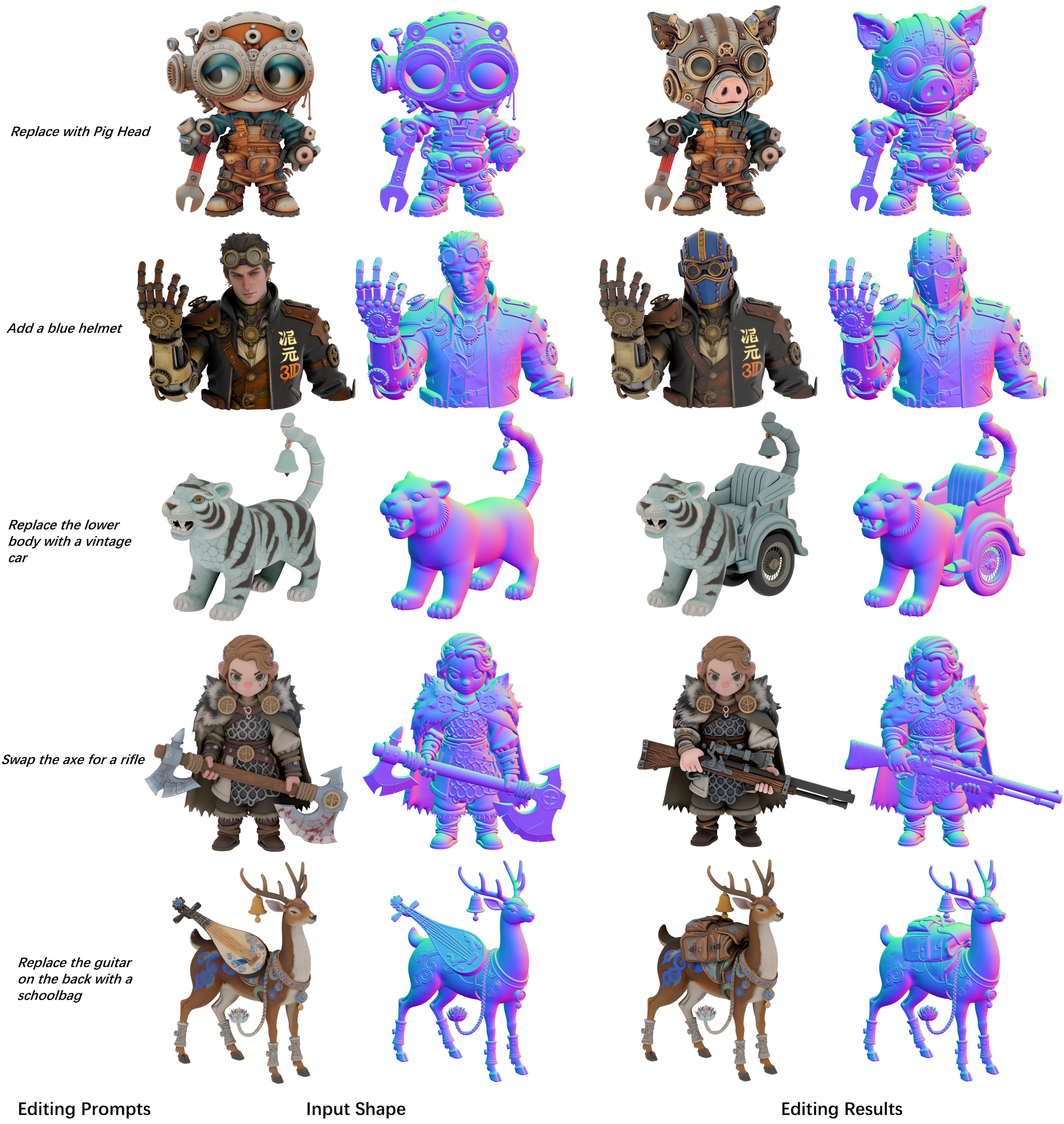}
  \caption{Examples of editing pairs in the training corpus created by Nano3D-v2.}
  \label{fig:editing_data_example}
\end{figure}

\begin{figure}[ht!]
  \centering
  \includegraphics[width=.9\linewidth]{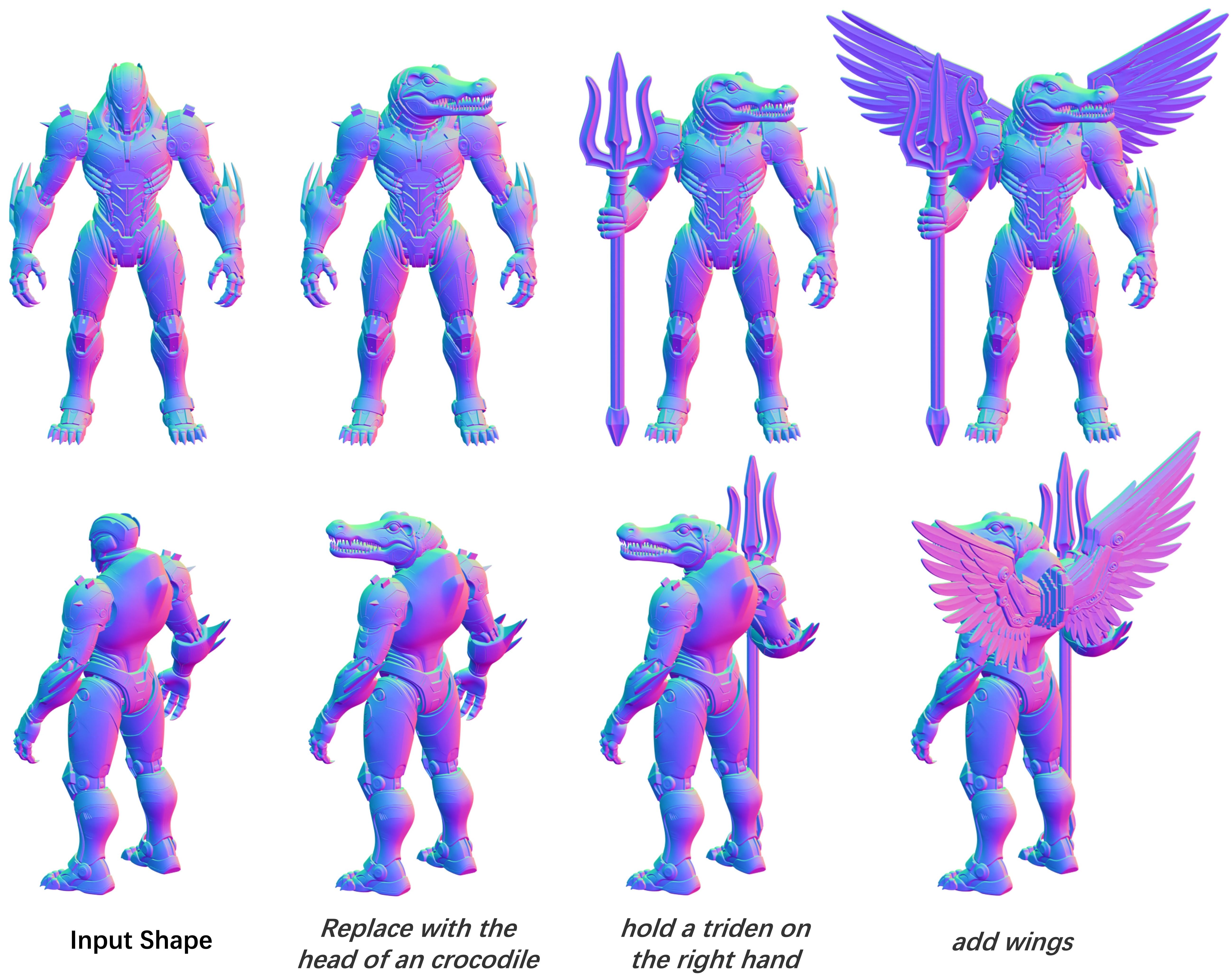}
  \caption{Examples of multi-round editing by Nano3D-v2.}
  \label{fig:multi_round_editing}
\end{figure}

This section details our automated pipeline for constructing high-fidelity 3D editing data. Given a source 3D asset and a natural-language instruction, the objective is to generate a target asset that faithfully executes the requested edit while preserving the geometry, identity, and multi-view consistency of non-target regions. To achieve this, we present \textbf{Nano3D-v2}, a comprehensive framework that integrates anchor-based viewpoint selection, learned 3D edit-region localization, voxel-level local modification, fine-grained geometry refinement, and multimodal quality filtering.

Current methods for constructing 3D editing data can be broadly categorized into two paradigms. The first follows a 2D-to-3D lifting strategy, where edits are first performed in the image domain and then lifted back into 3D space. For example, ShapeLLM-Omni~\cite{ye2025shapellm} leverages off-the-shelf 2D diffusion models to generate target visual conditions, followed by multi-view reconstruction. Although this paradigm benefits from the strong generative capability of 2D models, it lacks explicit 3D correspondences between the source asset and the edited target. As a result, the reconstructed 3D assets often suffer from identity drift, geometric hallucinations, multi-view inconsistency, and unintended modifications to regions that should remain unchanged.

The second paradigm introduces explicit 3D spatial constraints to regularize the editing process, as exemplified by Nano3D~\cite{ye2025nano3d} and VoxHammer~\cite{li2025voxhammer}. By restricting modifications to localized 3D regions, these methods improve spatial consistency and better preserve unedited areas. However, they are still limited in several aspects. First, the geometric quality of the edited results is largely bounded by the underlying 3D backbone, such as TRELLIS~\cite{xiang2025structured}, which often leads to over-smoothed or inaccurate local structures. Second, methods such as VoxHammer require manually specified 3D bounding boxes, making them difficult to scale to large-scale automatic data construction. Third, heuristic region estimation, as adopted in the original Nano3D~\cite{ye2025nano3d}, often fails to precisely localize small, occluded, or view-dependent edits.

Nano3D-v2 addresses these shortcomings by introducing a dedicated, learned 3D localization module and a high-fidelity refinement stage. As illustrated in Fig.~\ref{fig:nano3d-v2-pipeline}, the pipeline comprises three core modules—Editing Planning, Voxel Editing, and Detailed Geometry Refinement—together with rendering-based filtering and annotation, executed across 5 distinct stages.

\paragraph{Stage 1: Anchor View Selection.}
Starting with a source 3D asset, we render eight canonical views. Instead of arbitrary view selection, we employ a vision-language model to identify the optimal \textit{editing anchor}—the viewpoint that provides the highest semantic salience for the requested edit. We then utilize Qwen-Image to perform instruction-guided editing on this anchor view. This strategy ensures that the subsequent 3D optimization is grounded in the most informative 2D visual condition rather than a randomly selected perspective.

\paragraph{Stage 2: Editing Plaining.}
To surpass the limitations of heuristic masking, we develop a dedicated \textbf{Editing Planning} model. We first derive a 2D editing mask by computing the pixel-wise discrepancy between the source and edited anchor images. We then train an autoregressive Transformer to predict the corresponding 3D editing region. By feeding the 2D mask and the source voxel representation into this model, it learns to output a precise 3D bounding box. This box serves as a hard spatial constraint: voxels within the volume are mutable, while those outside remain frozen to preserve the original geometry.

\paragraph{Stage 3: Voxel Editing.}
Conditioned on the edited anchor image and the predicted 3D box, we employ the voxel Transformer from TRELLIS~\cite{xiang2025structured} to perform voxel-level FlowEdit. To further mitigate spurious deformations in unedited areas, we follow the local editing paradigm introduced in Nano3D~\cite{ye2025nano3d} by implementing a \textit{voxel-merge} operation. By replacing the edited voxels outside the predicted box with their original source counterparts, we explicitly enforce local editing and prevent global identity drift.

\paragraph{Stage 4: Fine-grained Geometry and Texture Editing.}
While voxel-level editing establishes a consistent global structure,
it lacks the resolution required for fine surface details and texture editing.
We therefore employ LATTICE~\cite{lai2025lattice} for sub-voxel geometry refinement.
Using the merged voxel as a geometric prior,
the LATTICE Transformer performs inversion-based inpainting.
This process refines the edited manifold while seamlessly stitching it with the source mesh,
resulting in high-resolution surfaces and artifact-free transitions around the editing boundary.
We further employ a native texture-generation model~\cite{lai2025natex} for texture editing. 
Given a source textured mesh, an edited mesh produced by LATTICE, an edited reference image, and a 3D edit region, the texture-generation transformer performs inversion-based texture inpainting. Because inpainting alone may still introduce discrepancies near the edit boundary, we additionally apply alpha blending over a spatial band around the boundary to ensure a smooth transition.

 
\paragraph{Stage 5: Edit Pairs Annotation and Filtering.}
The raw edit pairs produced by the preceding stages are noisy: some
edited meshes are structurally broken, and the original user prompts are
often coarse, ambiguous, or only loosely aligned with the realized
geometric change. To turn these pairs into reliable supervision, we apply
a vision-language-model-based annotation and filtering procedure that
operates on multi-view renders of the source and edited assets, covering
three aspects:

\begin{itemize}
  \item \textbf{Integrity filtering.} A VLM classifies each subject as a
    \emph{Human} or an \emph{Object} and scores its structural
    completeness under category-specific criteria, for instance applying
    stricter body and face requirements to humans while focusing on the
    presence and connectivity of major components for objects. Pairs
    whose edited asset fails to meet the integrity requirement are
    discarded.
  \item \textbf{Instruction annotation.} For the surviving pairs, we
    re-derive the editing instruction directly from the geometry rather
    than trusting the original prompt. The VLM compares the before and
    after assets, labels the edit as one of \emph{Addition},
    \emph{Replacement}, or \emph{Removal}, and produces a concise and a
    detailed instruction describing the most prominent geometric change
    first, constrained to geometry alone and ignoring color, material,
    and texture.
  \item \textbf{Verification.} We verify each annotated pair along four
    dimensions: the structural integrity of both assets, the alignment
    between the generated instruction and the actual visible change, the
    geometric consistency of the non-edited regions, and the visibility
    of the edit under the given viewpoints.
\end{itemize}

Overall, Nano3D-v2 harmonizes the semantic flexibility of 2D generative models with the structural rigor of learned 3D spatial constraints. By bypassing the pitfalls of direct lifting and manual intervention, our pipeline provides a scalable solution for generating consistent, high-quality 3D editing pairs.
Examples of editing pairs in the training corpus created by Nano3D-v2 are shown in Fig.~\ref{fig:editing_data_example}.
Multi-round editing is also supported by Nano3D-v2, as shown in Fig.~\ref{fig:multi_round_editing}.

\subsection{Text-grounded Part Generation Data}
\label{sec:data:part}

Vision Banana~\cite{gabeur2026visionbanana} shows that casting dense
perception (e.g., semantic, instance, and referring segmentation) as a
conditional \emph{generation} problem yields strong zero-shot transfer,
making image generation a \emph{universal interface} for heterogeneous
tasks. This motivates the same recipe in 3D: \emph{3D segmentation via
instruction tuning on a 3D multimodal large language model (MLLM)},
where part understanding is one instruction-following \emph{subtask}
prompted in natural language (e.g., ``segment the wheels'').

Unlike prior part generation
methods~\cite{ma2025p3,yan2025xpart,zhu2026cubepart}, which are
specialized pipelines relying on geometry-specific modules and
bounding-box supervision, we treat part understanding as a native
subtask of our unified 3D MLLM: we build instruction-tuning data
pairing 3D assets with natural-language queries and part-level targets,
so the same model used for understanding, generation, and editing can
also localize and decompose parts on demand. 

PartNeXt~\cite{wang2025partnext} provides part-level decompositions with
semantic grounding, where each segmented region is linked to a part
concept that can be referred to in language. Such labels can be directly
used to build instruction-following data, e.g., ``segment the wheels'' or
``remove the handle''. However, assets from datasets, such as
HY3D-Bench~\cite{hunyuan3d2026hy3dbenchgeneration3dassets}, only provide
raw mesh-level part decompositions. These parts are usually geometric
pieces created during modeling or export, and they do not have semantic
labels. They are also often over-segmented: one semantic part, such as a
wheel, wing, handle, or head, may be split into several disconnected
mesh components. As a result, each raw component is often too small or
ambiguous to be named by a clear semantic noun, making it difficult to
directly convert these decompositions into part-level instruction-tuning
data.

To address this issue, we design a \textbf{semantic mesh merging tool}
that turns over-segmented mesh components into semantic macro parts.
Here, a macro part means a coarse semantic part group: it may contain
multiple raw mesh components, but corresponds to one part concept that
can be referred to in language. The tool consists of three stages:
semantic part vocabulary discovery, component-to-part grounding and
merging, and quality filtering with instruction packaging.

\paragraph{Stage 1: Semantic Part Vocabulary Discovery.}
  For each object, we first render the full mesh from multiple canonical
  views and ask a VLM to infer a compact set of meaningful part concepts,
  such as wheels, wings, handles, or head. These concepts form an
  object-specific part vocabulary. We constrain the vocabulary to use
  simple and mutually exclusive names, so that each concept can be used
  directly in natural-language instructions.

\paragraph{Stage 2: Component-to-Part Grounding and Merging.}
  We then render each raw mesh component with a red highlight from multiple
  views. For each highlighted component, we provide the VLM with the
  object-specific part vocabulary obtained in Stage~1 and ask which
  semantic part the component belongs to. The model selects from this
  closed vocabulary, with an extra \emph{unmatched} option for unclear
  fragments. Components assigned to the same semantic concept are grouped
  together as one macro part. This step converts low-level geometric
  fragments into language-referable part targets.

\paragraph{Stage 3: Quality Filtering and Instruction Packaging.}
  Before using the merged parts as training data, we re-render each macro
  part as a red-highlighted region on the full object and use a VLM to
  verify its quality. The model checks whether the highlighted region
  matches the intended concept, whether the grouped components form a
  coherent part, and whether the part looks visually natural. We keep only
  clean object--part pairs that pass these checks. For each kept pair, we
  compute one normalization transform from the full object and apply the
  exact same scale and translation to all three exported assets: the full
  object, the macro part alone, and the remaining object after removing
  that macro part. These triplets are then converted into
  instruction-tuning samples for part segmentation and part removal, such
  as ``segment the wheels from this vehicle model'' and ``remove the
  wheels from this vehicle model''.


\section{Method}
\subsection{Architecture}
\label{sec:Architecture}

\begin{figure}[h!]
  \centering
  \includegraphics[width=\linewidth]{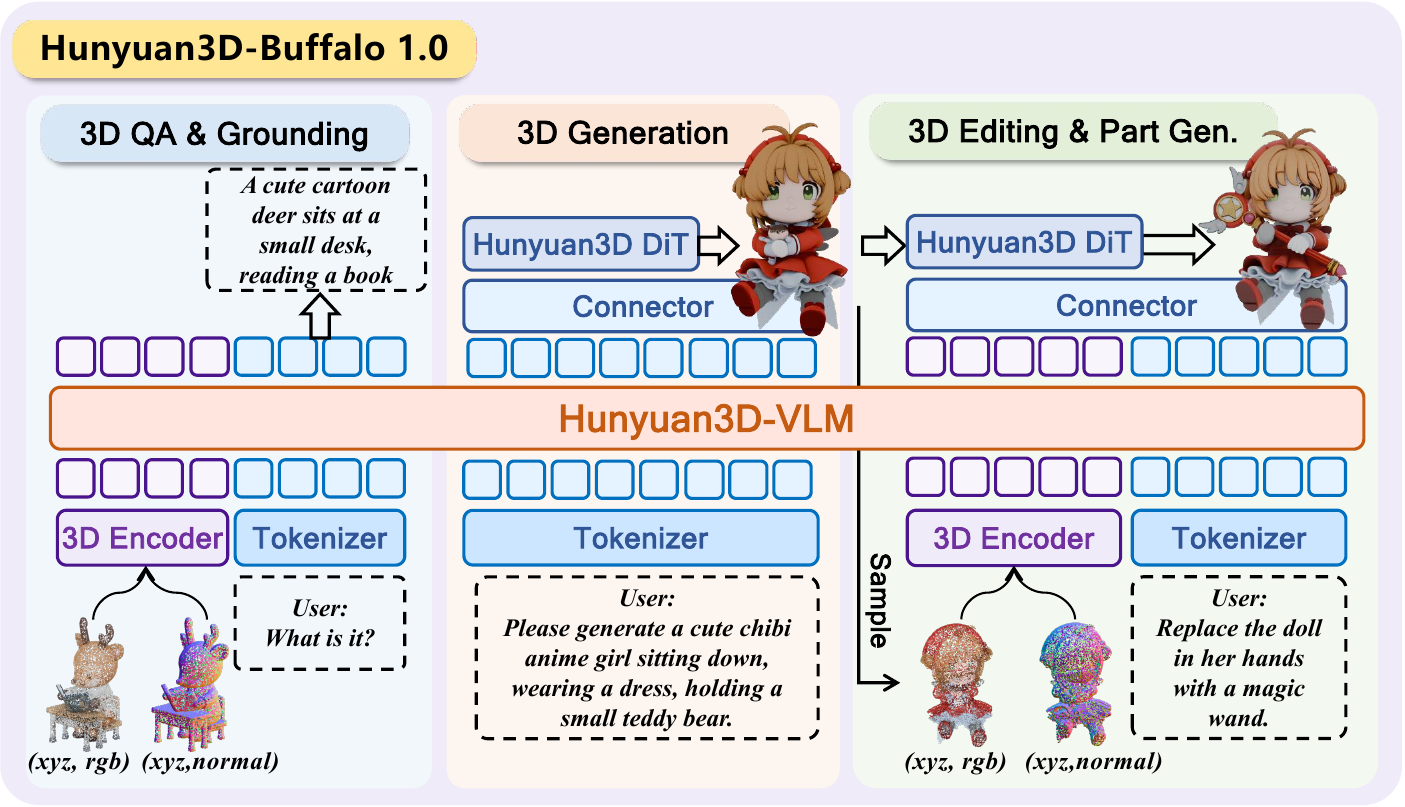}
  \vspace{-0.5cm}
  \caption{\textbf{Hunyuan3D-Buffalo 1.0 pipeline.} The framework unifies 3D QA and grounding, text-to-3D generation, and 3D editing through a shared Hunyuan3D-VLM backbone, which connects language, 3D representations, and generative Hunyuan3D DiT modules for multimodal understanding, generation, and editing.}
  \label{fig:pipeline}
\end{figure}

Inspired by hybrid frameworks such as Qwen-Image~\cite{wu2025qwen}, our architecture synergizes 
two specialized modules: (1) Hunyuan3D-VLM for multimodal understanding and 
part-level reasoning, and (2) 3D-DiT (initialized from Hunyuan3D-2.1) for 3D 
synthesis. While the VLM serves as the semantic engine, the 3D-DiT functions 
as the generative module. To integrate them, a lightweight MLP-Connector aligns 
VLM hidden states with the DiT's conditional space, ensuring high-level reasoning
 effectively guides the diffusion process without disrupting pretrained priors.

\paragraph{3D-aware Vision Language Model.}
To endow the VLM with fine-grained 3D perception, 
Hunyuan3D-VLM encodes 3D assets through a structure-and-appearance 
representation. Given a colored point cloud, the structural pathway 
processes geometric signals, including XYZ coordinates and surface 
normals, to capture object shape, spatial layout, and part boundaries. 
In parallel, the semantic pathway encodes RGB appearance cues, which 
helps the model distinguish parts that may be geometrically similar 
but visually different. The resulting 3D representations are encoded into 
latent tokens with a VecSet encoder. Before being injected into the VLM, these 
tokens are further compressed by a Q-Former into a fixed-length sequence of 
512 tokens, enabling efficient fusion with text and image tokens.

To support explicit 3D grounding and part-level reasoning, we further
augment the VLM vocabulary with 133 special tokens, following
Part-X-MLLM~\cite{wang2025part}. Three of them, \texttt{<|point\_start|>},
\texttt{<|point\_end|>}, and \texttt{<|point\_pad|>}, delimit the encoded
point-cloud token sequence and mark its placeholder positions within the
multimodal input. The remaining tokens encode 3D bounding boxes:
\texttt{<boxs>} and \texttt{<boxe>} act as box delimiters, while 128 discrete
coordinate tokens, from \texttt{<box-0>} to \texttt{<box-127>}, represent
quantized coordinate values in the range $[0, 127]$. Each 3D bounding box is
therefore represented as six quantized coordinate tokens wrapped by the box
delimiters. A decoder-only transformer, initialized from a pretrained
LLM, takes the fused sequence of structural, semantic, visual, and textual
tokens as input and autoregressively predicts the textual and coordinate-token
output. In this way, diverse 3D understanding tasks, including 3D captioning,
3D question answering, 3D grounding, edit-instruction synthesis, and
edit-outcome captioning, are unified as an instruction-following sequence
prediction problem.

\paragraph{Conditioning 3D-DiT with VLM Hidden States.}
For 3D generation, Hunyuan3D-VLM processes multimodal prompts—comprising interleaved text, images, and 3D shapes—to extract contextual hidden states. These high-level semantic features are then projected via the MLP-Connector into the 3D-DiT’s conditional embedding space. By initializing the 3D-DiT from Hunyuan3D-2.1, our architecture leverages robust 3D generative priors while utilizing the connector as a flexible interface for VLM-driven reasoning. This synergy effectively bridges the instruction-following and multimodal reasoning of an autoregressive VLM with the high-fidelity synthesis of a diffusion transformer.

\paragraph{3D Editing and Part Generation with Source-Object Conditioning.}
To ensure structural consistency during 3D editing and part generation, we condition the diffusion process on both VLM-derived semantic embeddings and the original object's representation. Specifically, the source 3D representation is concatenated with the noisy latent map as input to the 3D-DiT's self-attention layers. This explicit conditioning grants the model direct access to the original geometry during denoising, thereby facilitating the faithful preservation of unedited regions while enabling precise modifications to the target parts as instructed.
Note that 3D editing and part generation share the same data preparation and training pipeline: both require a source shape along with an instruction (either an editing instruction or a part segmentation instruction), and their data flows during training are identical.

\subsection{Training Procedure}

\begin{figure}[ht]
  \centering
  \resizebox{\linewidth}{!}{\begin{tikzpicture}[
    base/.style = {
        rectangle, 
        rounded corners=8pt, 
        draw=black, 
        minimum width=3cm, 
        minimum height=1.2cm, 
        text centered, 
        align=center,
        font=\sffamily\small,
        line width=0.8pt
    },
    vlm/.style = {base, fill=black!15}, 
    pre3d/.style = {base, fill=blue!10}, 
    omni/.style = {base, fill=indigo!10}, 
    edit/.style = {base, fill=yellow!20}, 
    sft/.style = {base, fill=pink!20}, 
    partgen/.style = {base, fill=green!15}, 
    arrow/.style = {-{Stealth[scale=1.2]}, line width=1pt, draw=gray!80},
    boxlabel/.style = {font=\sffamily\bfseries\scriptsize, text=gray!80, anchor=north west}
]

    \node (vlm) [vlm] {3D-VLM \\ Pre-training};
    \node (pre3d) [pre3d, right=0.8cm of vlm] {Text-to-3D \\ Pre-training};
    \node (omni) [omni, right=0.8cm of pre3d] {Omni \\ Pre-training};

    \draw [arrow] (vlm) -- (pre3d);
    \draw [arrow] (pre3d) -- (omni);

    \node (edit) [edit, above right=0.9cm and 1.4cm of omni] {3D Editing \\ Continued Pre-training};
    \node (sft) [sft, right=1.4cm of omni] {Text-to-3D \\ Continued Pre-training};
    \node (partgen) [partgen, below right=0.9cm and 1.4cm of omni] {Part-Gen \\ Continued Pre-training};

    \draw [arrow] (omni.east) -- ++(0.4cm,0) |- (edit.west);
    \draw [arrow] (omni.east) -- (sft.west);
    \draw [arrow] (omni.east) -- ++(0.4cm,0) |- (partgen.west);

\end{tikzpicture}}
  \caption{Overview of the training pipeline. The shared trunk proceeds through 3D-VLM pre-training, text-to-3D pre-training, and omni pre-training. It then branches into three continued pre-training paths: 3D editing, text-to-3D, and part generation.}
  \label{fig:training_pipeline}
\end{figure}
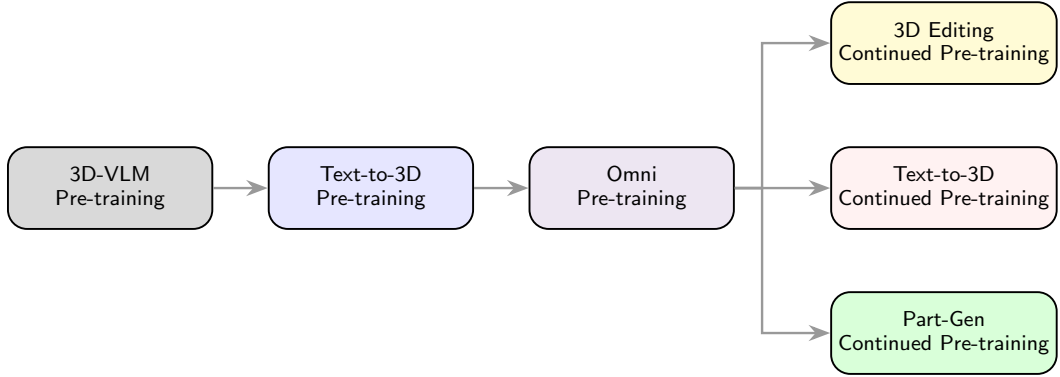

As illustrated in Fig.~\ref{fig:training_pipeline}, we train our framework in
four stages: 3D-VLM pre-training (Stage~1), text-to-3D pre-training (Stage~2),
omni pre-training (Stage~3), and continued pre-training (Stage~4) where the model branches into three task-specific paths for 3D editing, text-to-3D, and part generation respectively. All
generative stages optimize the 3D-DiT with a flow-matching objective that
predicts the velocity field transporting a Gaussian prior to the target 3D
latent distribution.

\paragraph{Stage 1: 3D-VLM Training.}
We first endow Hunyuan3D-VLM with fine-grained 3D perception in two phases.
In the \emph{alignment} phase, we freeze the Qwen-VL\cite{bai2025qwen2} backbone and train only
the 3D-token connector (VecSet encoder, Q-Former, and projection) to align
3D latent tokens with the language model's input embedding space. In the
\emph{instruction-tuning} phase, we unfreeze the full model and jointly train
it on text-to-text, image-to-text, 3D grounding, 3D captioning, 3D question
answering, and edit-instruction synthesis. The resulting VLM serves as a
unified semantic engine and is kept frozen in all subsequent stages, so that
generative training focuses on the 3D-DiT and the connector without
disturbing the learned 3D understanding.

\paragraph{Stage 2: Text-to-3D Pretraining.}
We then couple the VLM with the 3D-DiT for text-conditioned generation. 
The VLM hidden states are first processed by the MLP-Connector and then injected 
into the 3D-DiT, where the resulting per-token features condition the denoiser 
through cross-attention. Initialized from Hunyuan3D-2.1, the
3D-DiT is trained on our $\sim$50M text--asset pairs, 
establishing broad coverage of categories, structures, and compositions as a
strong initialization for both branches.

\paragraph{Stage 3: Omni Pretraining.}
We pretrain the model on text-to-3D, 3D editing, and part generation tasks in a unified manner.
Overall, the sampling ratio is set to text-to-3D : (3D editing + part generation) = $1:1$.
This balance is essential: it lets the editing capability emerge without
sacrificing generation, which in turn underpins editing that generalizes
beyond the training distribution.  
Since the editing and part generation corpus is far smaller than the text-to-3D one, we repeat the editing and part generation data by $4\times$ to match the text-to-3D data within each batch.
Part generation is regarded as a special case of 3D editing, where the target region is the queried part itself.

\paragraph{Stage 4: Continued Pre-training.}
In the final stage, we decouple the three tasks and train them separately.
For 3D editing and part generation, we still mix in half text-to-3D data during training to preserve generative quality.
For text-to-3D, we train exclusively on text-to-3D data.

\section{Experiments}

\subsection{3D Understanding}

\paragraph{Evaluation setting.}
We evaluate the 3D understanding capability of Hunyuan3D-VLM on UniPart-Bench~\cite{ye2025shapellm}, 
a part-centric benchmark for structured 3D perception and language understanding. 
Each 3D object is represented as an RGB point cloud and annotated with axis-aligned 
part bounding boxes, part-level texts, object-level captions, and part-aware 
question-answer pairs. Following the benchmark setting, part annotations are 
provided at two semantic granularities: Q1 denotes the coarse part category or 
name, while Q2 denotes a fine-grained natural-language description of the part. 
For comparison, we include representative 3D multimodal baselines,
including GPT4Point~\cite{qi2024gpt4point}, PointLLM~\cite{xu2024pointllm},
ShapeLLM~\cite{qi2024shapellm}, ShapeLLM-Omni~\cite{ye2025shapellm},
Part-X-MLLM~\cite{wang2025part}, and UniVerse3D~\cite{ye2026universe3d}.

\paragraph{Tasks and metrics.}
The all-task evaluation covers the understanding-related tasks in UniPart-Bench. 
Task 0, \emph{Pure box listing}, requires the model to predict all part bounding 
boxes without text generation. Tasks 1--2 are \emph{Multi-Part Grounding}, where 
the model outputs all part boxes together with Q1 labels or Q2 descriptions. 
Tasks 3--4 are \emph{Single-Part Grounding}, where the model localizes a queried
part specified by either a Q1 part name or a Q2 fine-grained description. 
Tasks 5--6 are \emph{Box-to-Text}, where the model generates the corresponding 
Q1 label or Q2 description given a part box. Task 7 is \emph{Part QA}, which 
evaluates part-level question answering with grounding. We report IoU for bounding-box
localization and use SBERT, SimCSE, BLEU-1, ROUGE-L, and METEOR for textual outputs.

\vspace{-0.7cm}
\begin{table*}[h!]
\centering
\caption{\textbf{Comparison of part-level question answering and object-level captioning on UniPart-Bench~\cite{ye2025shapellm}.}
The two tasks assess complementary aspects of 3D understanding, including localized part-aware reasoning and holistic object-level semantic description.}
\label{tab:unipart_qa_captioning}
\resizebox{\linewidth}{!}{
\begin{tabular}{l|ccccc|ccccc}
\toprule
\multirow{2}{*}{Model}
& \multicolumn{5}{c|}{Part Understanding Q\&A}
& \multicolumn{5}{c}{Overall 3D Object Captioning} \\
\cmidrule(lr){2-6} \cmidrule(lr){7-11}
& SBERT & SimCSE & BLEU-1 & ROUGE-L & METEOR
& SBERT & SimCSE & BLEU-1 & ROUGE-L & METEOR \\
\midrule
GPT4Point~\cite{qi2024gpt4point}
& 48.32 & 45.17 & 15.16 & 22.55 & 16.19
& 25.60 & 27.00 & 11.50 & 12.00 & 12.70 \\
PointLLM-7B~\cite{xu2024pointllm}
& 61.30 & 58.48 & 21.78 & 29.26 & 22.45
& 42.79 & 42.44 & 11.58 & 14.39 & 16.90 \\
PointLLM-13B~\cite{xu2024pointllm}
& 56.36 & 51.47 & 21.40 & 29.16 & 21.80
& 43.51 & 43.12 & 13.54 & 15.74 & 17.45 \\
ShapeLLM-13B~\cite{qi2024shapellm}
& 61.19 & 57.26 & 23.32 & 32.56 & 24.45
& 25.15 & 27.14 & 11.77 & 12.14 & 12.84 \\
ShapeLLM-Omni-7B~\cite{ye2025shapellm}
& 57.35 & 51.16 & 22.77 & 29.57 & 23.24
& 31.18 & 31.93 & 17.79 & 19.04 & 14.30 \\
Part-X-MLLM~\cite{wang2025part}
& 78.98 & 84.25 & 40.54 & 42.26 & 34.24
& 53.82 & 51.97 & 36.04 & 38.11 & 30.71 \\
UniVerse3D~\cite{ye2026universe3d}
& 83.11 & 87.16 & 46.79 & 43.94 & 42.05
& 65.18 & 66.25 & 42.75 & 44.17 & 41.11\\
\midrule
\rowcolor{gray!12}
\textbf{Hunyuan3D-VLM (Ours)}
& \textbf{85.47} & \textbf{89.06} & \textbf{49.95} & \textbf{45.01} & \textbf{45.79}
& \textbf{72.94} & \textbf{73.60} & \textbf{50.93} & \textbf{52.84} & \textbf{50.47} \\
\bottomrule
\end{tabular}
}
\end{table*}

\vspace{-0.8cm}
\begin{table*}[h!]
\centering
\caption{\textbf{Detailed all-task evaluation of Hunyuan3D-VLM on UniPart-Bench~\cite{ye2025shapellm}.}
The benchmark covers pure box listing, multi-part grounding, single-part grounding, box-to-text generation, and part-level question answering.}
\label{tab:unipart_all_task_results}
\resizebox{0.9\linewidth}{!}{
\begin{tabular}{clcccccc}
\toprule
Task & Name & IoU & SBERT & SimCSE & BLEU-1 & ROUGE-L & METEOR \\
\midrule
0  & Pure box listing              & 0.864 &  -    &   -   &  -    &   -   &  -    \\
1  & Multi-Part Grounding (Q1)     & 0.880 & 68.00 & 68.55 & 52.06 & 52.09 & 26.35 \\
2  & Multi-Part Grounding (Q2)     & 0.844 & 70.92 & 69.47 & 40.04 & 41.86 & 38.19 \\
3  & Single-Part Grounding (Q1)    & 0.626 & 78.95 & 77.92 & 45.74 & 47.47 & 44.07 \\
4  & Single-Part Grounding (Q2)    & 0.525 &  -    &  -    &   -   &  -    &   -   \\
5  & Box-to-Text (Q1)              &  -    & 67.64 & 68.27 & 49.89 & 50.00 & 25.45 \\
6  & Box-to-Text (Q2)              &  -    & 74.13 & 72.99 & 42.01 & 44.28 & 40.96 \\
7  & Part QA                       & 0.633 & 85.47 & 89.06 & 49.95 & 45.01 & 45.79 \\
\bottomrule
\end{tabular}
}
\end{table*}

\paragraph{Results.}
As shown in Table~\ref{tab:unipart_qa_captioning}, 
Hunyuan3D-VLM consistently achieves the best performance on both part-level Q\&A and overall 
3D object captioning. For part understanding Q\&A, our model obtains 85.47 SBERT and 89.06 SimCSE, 
with notably stronger lexical scores as well, including 49.95 BLEU-1 and 45.79 METEOR. These 
results indicate that Hunyuan3D-VLM can produce semantically accurate and linguistically faithful 
answers for fine-grained part-centric questions. For holistic object captioning, our model further 
reaches 72.94 SBERT and 52.84 ROUGE-L, showing a clear advantage over previous 3D 
MLLMs in object-level semantic description. Overall, the results demonstrate that Hunyuan3D-VLM 
improves 3D understanding at both localized part level and global object level.

Table~\ref{tab:unipart_all_task_results} further reports the detailed performance across 
UniPart-Bench tasks. Hunyuan3D-VLM obtains 0.864 IoU on pure box listing and achieves strong
grounding performance on both multi-part and single-part settings. In addition, the model 
produces competitive language scores for box-to-text generation and part QA, showing its 
ability to unify part localization, region-conditioned description, and part-aware reasoning 
within a single 3D understanding framework.

\begin{figure}[ht!]
  \centering
  \includegraphics[width=0.97\linewidth]{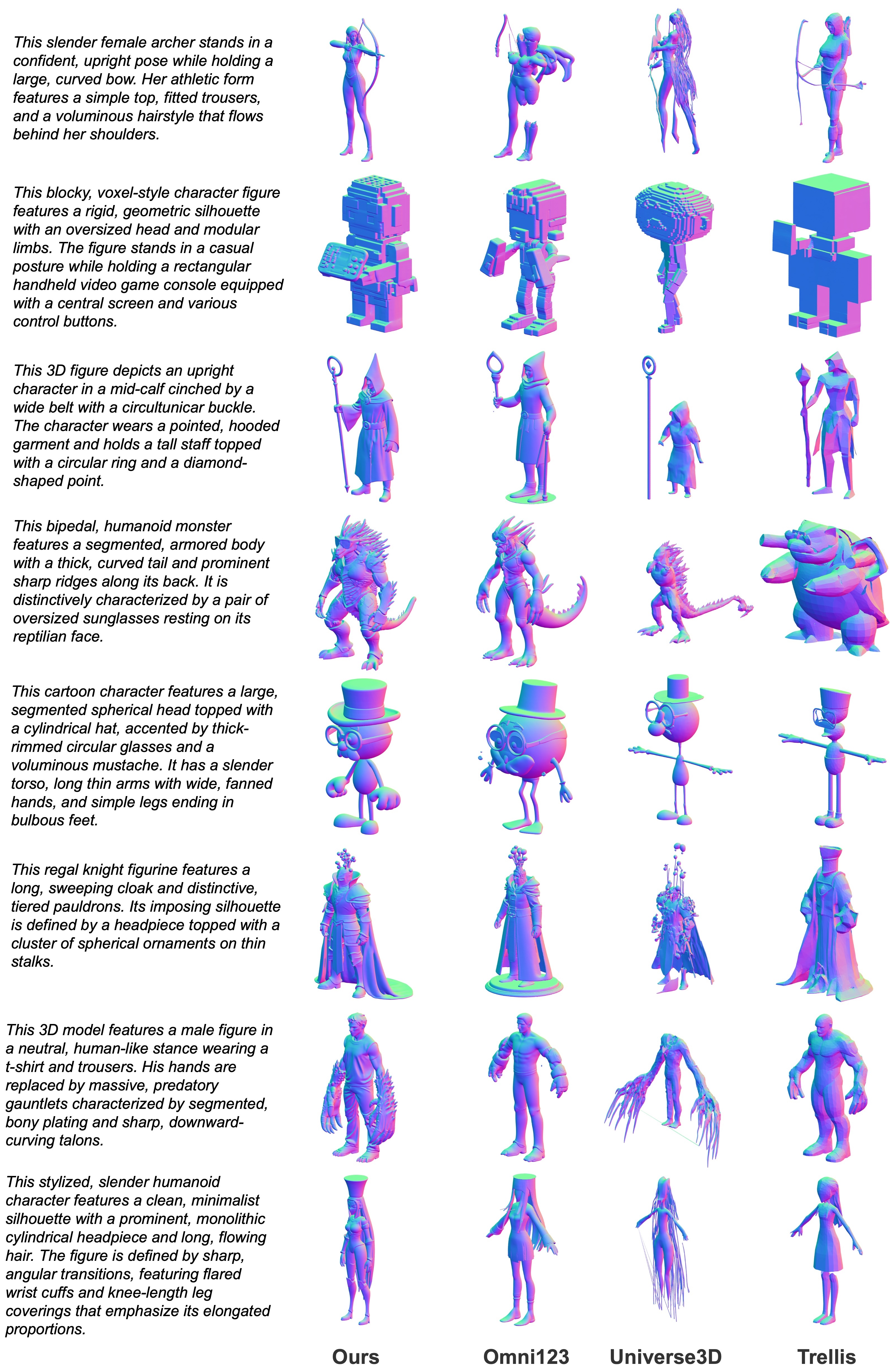}

  \caption{Qualitative text to 3D results.}
  \label{fig:t23d-compare1}
\end{figure}

\subsection{Text-to-3D} 
We conduct a human evaluation to compare our method against three baselines (TRELLIS~\cite{xiang2025structured}, Universe3D~\cite{ye2026universe3d}, and Omni123~\cite{ye2025omni123}). All methods are evaluated on the same set of 100 text prompts, producing 100 four-way comparison groups. To ensure the evaluation is unbiased and representative, the prompts are deliberately diverse in both content and length: they span a wide range of object categories—including characters and creatures, everyday objects, weapons and tools, furniture and decorative items, and clothing—and vary from short, concise descriptions (as few as 2 words, e.g., "Sitting cat.") to long, fine-grained specifications (up to 54 words), with an average length of roughly 28 words. About one fifth of the prompts are short tags, while the remainder are medium-to-long detailed descriptions, so that the comparison reflects performance across both coarse and highly specific inputs. For each prompt, the four generated 3D assets are presented anonymously and in randomized order to avoid position and brand bias. Participants are asked to choose the best result along three dimensions: (i) text alignment, i.e. how faithfully the asset matches the input prompt; (ii) geometry quality, i.e. the quality and plausibility of the 3D shape; and (iii) overall preference. For each method and dimension, we report the proportion of comparisons in which it was preferred.

\vspace{-0.7cm}
\begin{table*}[h!]
\centering
\caption{Results of the user study for text-to-3D generation. We report the
\emph{preference rate} (\%), i.e., the percentage of comparisons in which each
model is selected as the best among the four candidates. In each comparison,
the four results are shown side by side in randomized order, and participants
pick the single best result for each criterion (a ``tie'' option is also allowed, so
columns may sum to slightly below 100\%). Higher is better, with a random-choice baseline of 25\%. Best results are
shown in \textbf{bold}.}
\label{tab:user_study_t23d}
\resizebox{0.8\linewidth}{!}{
\begin{tabular}{l|ccc}
\toprule
Model & Text alignment & Geometry quality & Overall preference \\
\midrule
Universe3D~\cite{ye2026universe3d} & 8.2 & 7.4 & 8.3 \\
TRELLIS~\cite{xiang2025structured} & 14.9 & 12.4 & 14.4 \\
Omni123~\cite{ye2025omni123} & 17.5 & 21.0 & 18.4 \\
\midrule
\rowcolor{gray!12}
\textbf{Hunyuan3D-Buffalo 1.0 (Ours)} & \textbf{55.2} & \textbf{57.1} & \textbf{56.6} \\
\bottomrule
\end{tabular}
}
\end{table*}

\vspace{-0.7cm}
\begin{table*}[h!]
    \centering
    \caption{Ablation on the quantity of training data. We report the
    \emph{preference rate} (\%), i.e., the percentage of comparisons in which a
    model is chosen as the best among the three variants. In each comparison, the
    three results are shown side by side in randomized order, and participants
    select the best result for each criterion (a ``tie'' option is also allowed, so
    columns may sum to slightly below 100\%). Higher is better, and the
    random-choice baseline is 33.3\%. Best results are shown in \textbf{bold}.}
    \label{tab:t23d_data_ablation}
    \resizebox{0.8\linewidth}{!}{
    \begin{tabular}{l|ccc}
    \toprule
    Num. of samples & Text alignment & Geometry quality & Overall preference \\
    \midrule
    300w  & 9.9 & 8.8 & 8.4 \\
    1500w & 28.8 & 29.0 & 28.6 \\
    5000w & \textbf{54.5} & \textbf{57.4} & \textbf{57.5} \\
    \bottomrule
    \end{tabular}
    }
\end{table*}

As shown in Table~\ref{tab:user_study_t23d}, our method is preferred in the clear majority of comparisons across all three dimensions, attaining preference rates of 55.2\%, 57.1\%, and 56.6\% for text alignment, geometry quality, and overall preference, respectively. These rates are more than double those of the strongest baseline, Omni123~\cite{ye2025omni123} (17.5\%, 21.0\%, and 18.4\%), and far exceed the 25\% random-choice level, while TRELLIS~\cite{xiang2025structured} and Universe3D~\cite{ye2026universe3d} trail further behind. The margin is largest on geometry quality, indicating that our method produces noticeably more faithful and higher-fidelity 3D shapes, and it remains the most preferred for text alignment, showing that this geometric gain does not come at the cost of prompt faithfulness. Notably, the ranking is consistent across every individual participant and holds for both the short tag-style prompts and the longer fine-grained descriptions, suggesting that the advantage is robust to the diversity of input content and length rather than driven by a particular subset of cases.


\paragraph{Ablation of different quantity of training data}

Table~\ref{tab:t23d_data_ablation} ablates the effect of pretraining data scale, comparing models pretrained on 300w (3M), 1500w (15M), and 5000w (50M) samples. We observe a clear and monotonic improvement in human preference as the pretraining corpus grows: the preference rate for overall quality rises from $8.4\%$ (300w) to $28.6\%$ (1500w) and to $57.5\%$ (5000w), with the same trend holding for text alignment ($9.9\% \rightarrow 28.8\% \rightarrow 54.5\%$) and geometry quality ($8.8\% \rightarrow 29.0\% \rightarrow 57.4\%$). Notably, the model trained on the full 5000w data is preferred in well over half of all comparisons---roughly twice as often as the 1500w model and more than six times as often as the 300w model---and this ordering is consistent across every participant and all three evaluation criteria. The gains are pronounced on both text alignment and geometry quality, indicating that larger-scale pretraining improves not only how faithfully the generated 3D assets follow the input prompt but also the fidelity and plausibility of the underlying geometry. These results highlight that scaling up pretraining data is a key driver of generation quality, and that performance has not yet saturated, suggesting that further data scaling may yield additional improvements.


\begin{figure}[htbp]
  \centering
  \includegraphics[width=\linewidth]{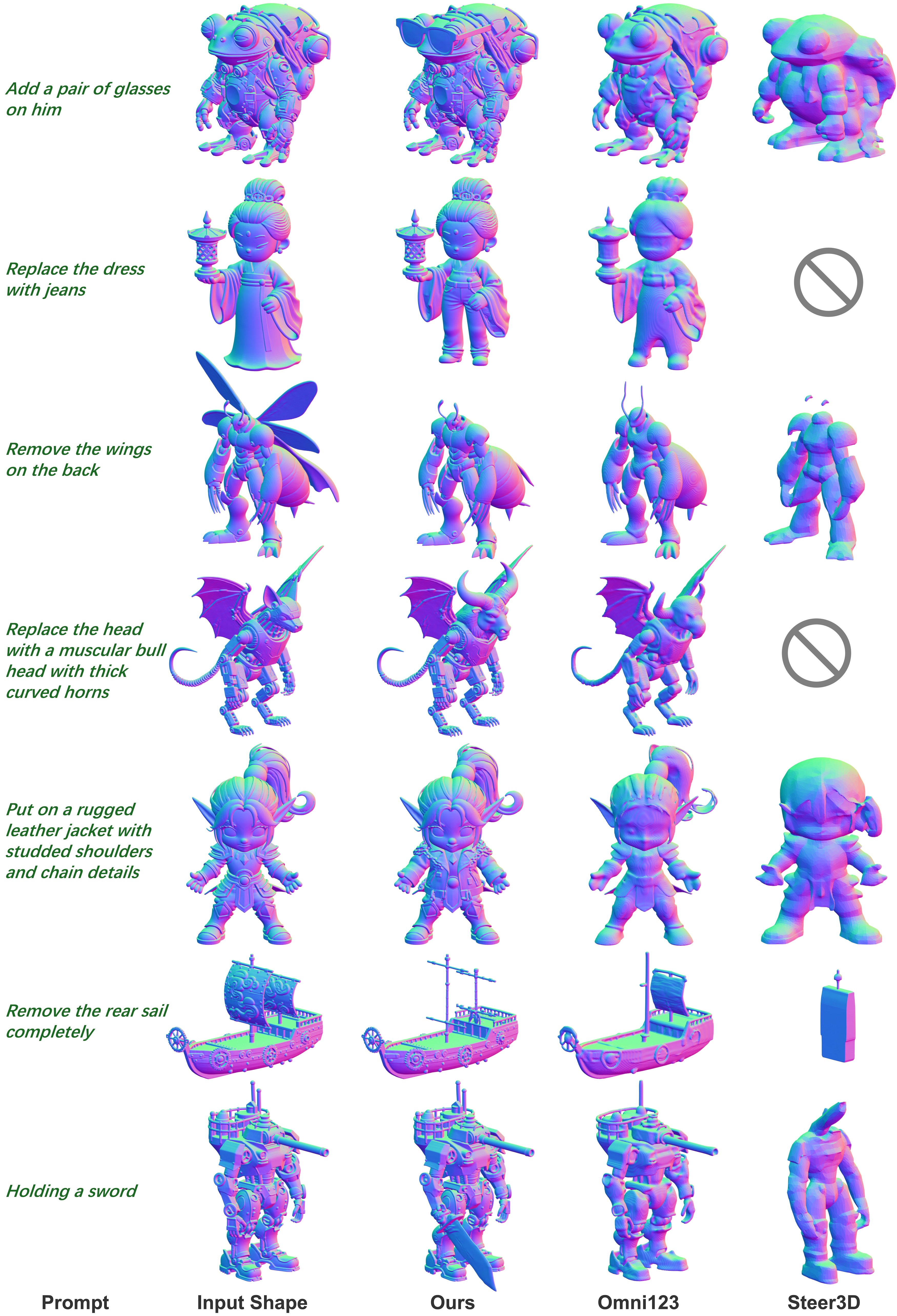}
  \caption{\textbf{Qualitative shape editing results.}
  Our method significantly outperforms all baselines in both geometric consistency before and after editing, and responsiveness to editing instructions.
  }
  \label{fig:edit_compare}
\end{figure}

\begin{figure}[h]
  \centering
  \includegraphics[width=\linewidth]{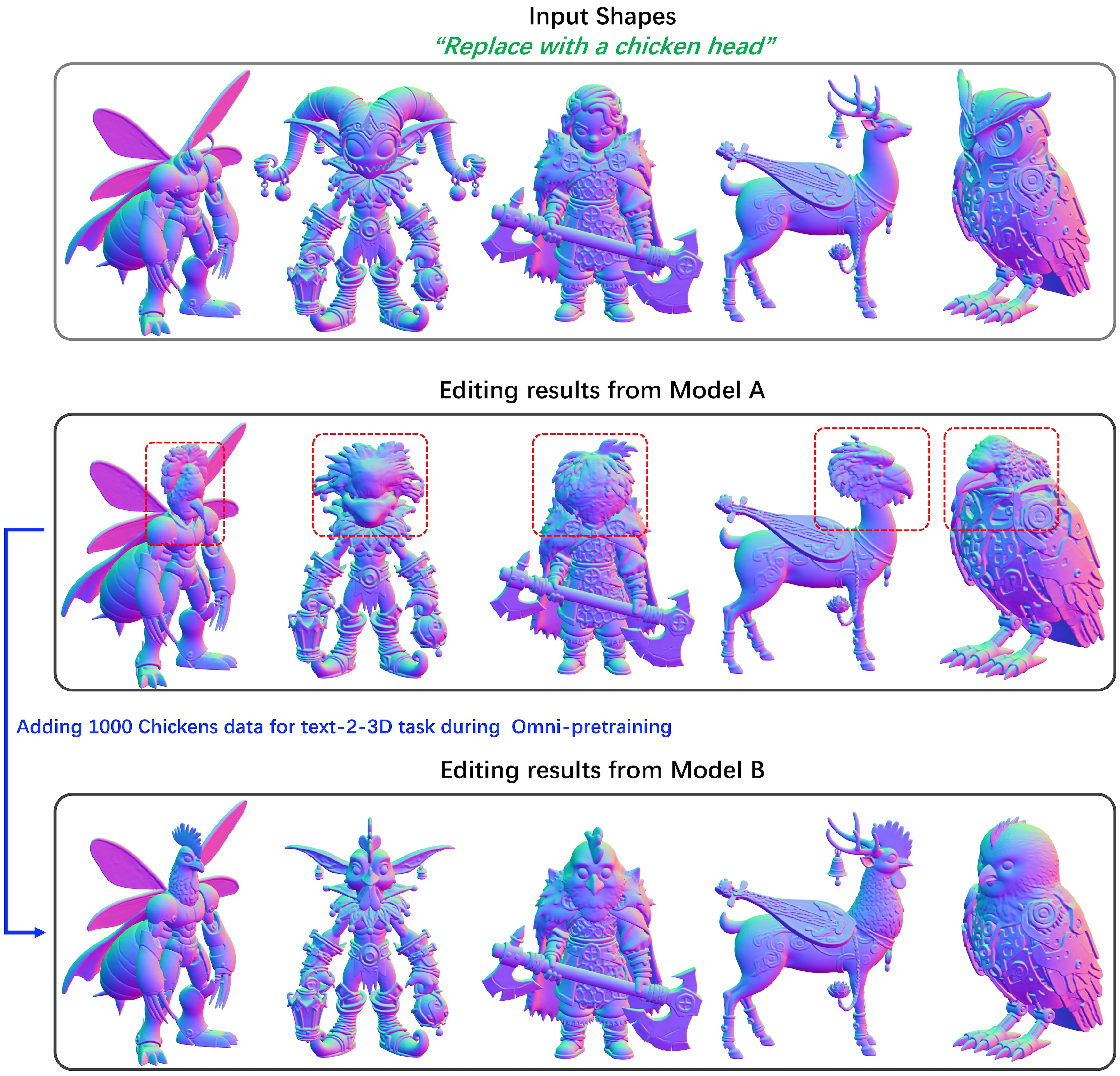}
  \caption{\textbf{Scaling up text-to-3d data facilitates 3D editing.}
  Model A is our base model; when instructed to
  edit an object by replacing its head with a chicken head, it fails to produce a
  satisfactory result.
  Model B is built upon Model A by adding only 1{,}000 additional
  chicken samples for the text-to-3D task during the \textbf{Omni pre-training stage}---crucially, without introducing any new editing data.
  After incorporating this text-to-3D data, the model can successfully replace the head with a chicken head.
  This suggests a clear direction: to improve 3D editing, the text-to-3D generation capability should be maximized as much as possible.
  Since constructing text-to-3D data is far less costly than constructing 3D editing data, scaling up text-to-3D data is a relatively more feasible path toward stronger 3D editing.
  }
  \label{fig:gen_help_edit}
\end{figure}

\subsection{3D Editing}

As the most direct and natural interface for 3D creation, language instruction-driven
3D editing is a pivotal capability for a wide range of 3D applications. We evaluate our
model on Edit3D-Bench~\cite{ma2025steer3d}, using its curated source-target mesh pairs for geometric addition
and removal operations. The objective is to perform semantically faithful and
structurally coherent modifications that closely follow the language instruction, while
keeping the unedited regions of the source mesh as intact as possible.

To quantify editing fidelity, we report Chamfer Distance (CD) and F1 score (F1) between
the generated and ground-truth edited meshes. Each test sample requires the model to
translate a natural-language directive into a precise, localized geometric transformation.
We compare against five recent 3D editing methods: ShapeLLM-Omni~\cite{ye2025shapellm},
Steer3D~\cite{ma2025steer3d}, 3DEditFormer~\cite{xia2025editformer},
Tailor3D~\cite{qi2024tailor3dcustomized3dassets}, and Omni123~\cite{ye2025omni123}. This
selection spans both native autoregressive editors and optimization-based pipelines.

\begin{table*}[t]
\centering
\caption{\textbf{Quantitative comparison of language-guided 3D shape editing on Edit3D-Bench~\cite{ma2025steer3d}.}
Both CLIP-conditioned and 3D-VLM-conditioned variants of Hunyuan3D-Buffalo 1.0 are included to analyze the effect of stronger 3D instruction understanding.}
\label{tab:edit3d_bench}
\setlength{\tabcolsep}{6pt}
\renewcommand{\arraystretch}{1.12}
\resizebox{\linewidth}{!}{
\begin{tabular}{lcccccc}
\toprule
\multirow{2}{*}{Method}
& \multicolumn{2}{c}{Add}
& \multicolumn{2}{c}{Remove}
& \multicolumn{2}{c}{Avg} \\
\cmidrule(lr){2-3}
\cmidrule(lr){4-5}
\cmidrule(lr){6-7}
& CD $\downarrow$ & F1 $\uparrow$
& CD $\downarrow$ & F1 $\uparrow$
& CD $\downarrow$ & F1 $\uparrow$ \\
\midrule
ShapeLLM-Omni~\cite{ye2025shapellm}
& 0.2546 & 0.0877
& 0.2237 & 0.1166
& 0.2392 & 0.1022 \\
3DEditFormer~\cite{xia2025editformer}
& 0.1676 & 0.1955
& 0.1342 & 0.1836
& 0.1509 & 0.1896 \\
Tailor3D~\cite{qi2024tailor3dcustomized3dassets}
& 0.1661 & 0.1217
& 0.1755 & 0.1352
& 0.1708 & 0.1285 \\
Steer3D~\cite{ma2025steer3d}
& 0.1404 & 0.2414
& 0.0976 & 0.3044
& 0.1190 & 0.2729 \\
Omni123~\cite{ye2025omni123}
& 0.0736 & 0.1743
& 0.0632 & 0.2259
& 0.0684 & 0.2001 \\
\midrule
\textbf{Hunyuan3D-Buffalo 1.0 w/ CLIP (Ours)}
& 0.0154 & \textbf{0.5657}
& 0.0162 & 0.7015
& 0.0158 & 0.6336 \\
\rowcolor{gray!12}
\textbf{Hunyuan3D-Buffalo 1.0 w/ 3D-VLM (Ours)}
& \textbf{0.0127} & 0.5610
& \textbf{0.0054} & \textbf{0.7420}
& \textbf{0.0091} & \textbf{0.6515} \\
\bottomrule
\end{tabular}
}
\end{table*}

\paragraph{Quantitative Comparison on Edit3D-Bench.}
Table~\ref{tab:edit3d_bench} reports the quantitative results on add and remove
editing tasks. Hunyuan3D-Buffalo 1.0 significantly outperforms previous 3D editing methods
across both CD and F1 metrics. Compared with the strongest baseline in terms of
average CD, Omni123~\cite{ye2025omni123}, our 3D-VLM version reduces the average
CD from 0.0684 to 0.0091, corresponding to an 86.7\% relative reduction. For
average F1, our model improves over the strongest baseline Steer3D~\cite{ma2025steer3d}
from 0.2729 to 0.6515, achieving a 2.39$\times$ improvement.
These improvements are consistent across both addition and removal tasks, 
where Hunyuan3D-Buffalo 1.0 w/ 3D-VLM achieves 0.0127/0.5610 and 0.0054/0.7420 in CD/F1, 
respectively, demonstrating accurate localized editing and strong preservation of 
the remaining geometry.

We also compare two conditioning variants of Hunyuan3D-Buffalo 1.0: the CLIP-based version
and the 3D-VLM-based version. The 3D-VLM version achieves a lower average CD than
the CLIP version, reducing it from 0.0158 to 0.0091, while improving the average F1
from 0.6336 to 0.6515. This indicates that the 3D-VLM-conditioned model follows
editing instructions more accurately and produces finer geometric details.

\paragraph{Qualitative Analysis of Localized and Consistent Editing.}
As shown in Fig.~\ref{fig:edit_compare}, our method performs localized geometric
edits while preserving the overall structure, pose, and fine-grained details of the
input shape. For addition tasks, such as adding glasses, a rugged leather jacket, or
a sword, our model introduces the target component at the correct semantic location
without disrupting unrelated regions. For removal tasks, such as removing the rear
sail or the wings on the back, the edited results retain the main body geometry and
avoid excessive deformation. In contrast, Omni123~\cite{ye2025omni123} often produces
over-smoothed shapes or incomplete edits, losing detailed geometry in the source mesh.
Steer3D~\cite{ma2025steer3d} shows less stable behavior: in several cases, it either
fails to produce a valid result or changes the global structure of the object instead
of applying a localized edit. These qualitative results indicate that our model better
balances instruction following and source-shape preservation, which is essential for
practical 3D editing.

\paragraph{Scaling Text-to-3D Data for Better Shape Editing.}
To better understand the relationship between 3D editing and 3D generation, we conduct
a deeper investigation into how the two interact. We find that 3D editing fundamentally
relies on a strong text-to-3D generative foundation: when the underlying text-to-3D
generation is weak, it is difficult to achieve strong editing capability.
Figure~\ref{fig:gen_help_edit} illustrates this with a concrete example. By adding more
text-to-3D training data, the corresponding editing ability simultaneously emerges
without additional editing data. This suggests a clear direction: to improve 3D editing,
the text-to-3D generation capability should be maximized as much as possible. Since
constructing text-to-3D data is far less costly than constructing 3D editing data,
scaling up text-to-3D data is a relatively more feasible path toward stronger 3D editing.

\subsection {Part Generation}

\begin{figure}[h!]
  \centering
  \includegraphics[width=1\linewidth]{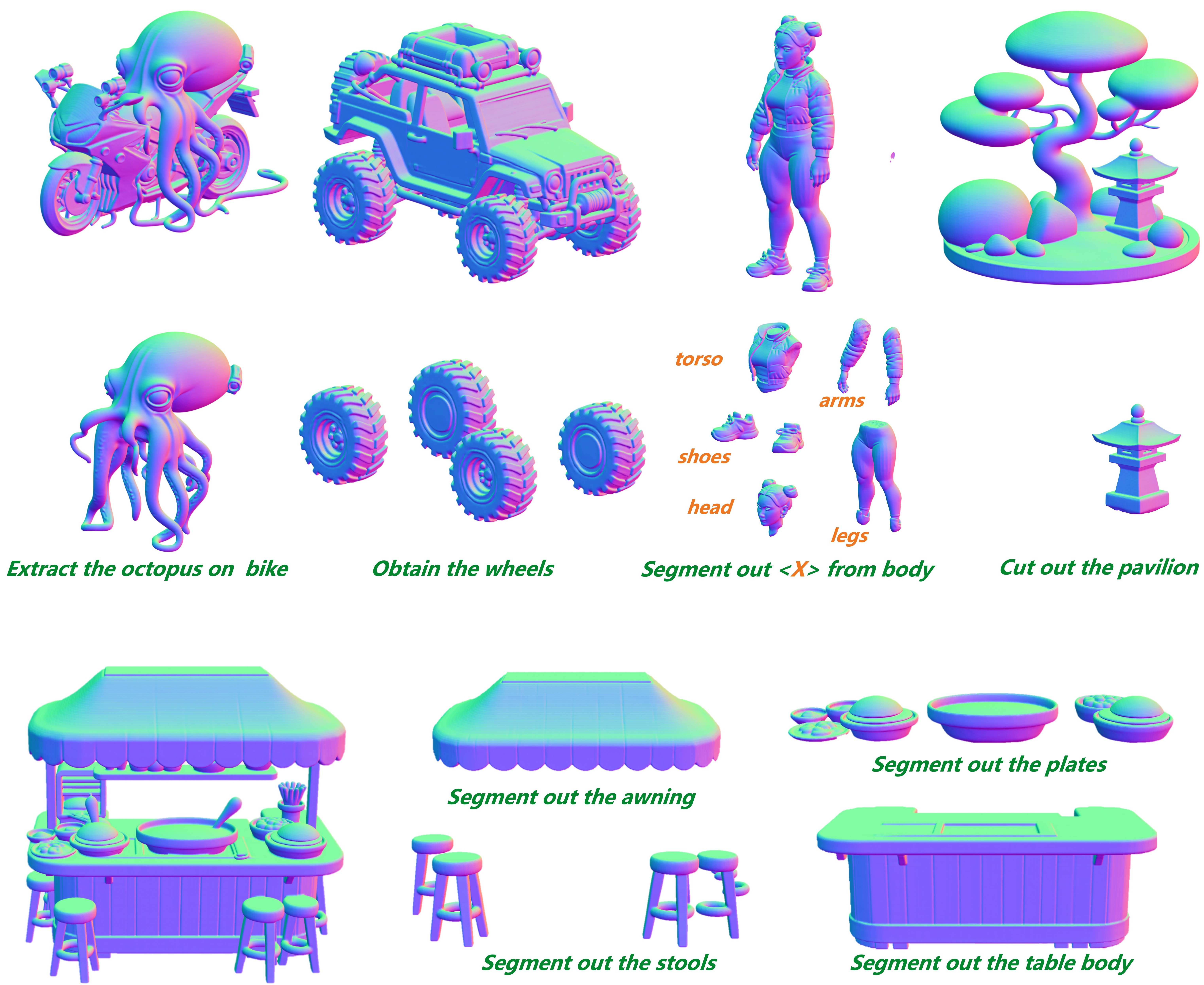}
  \caption{\textbf{Qualitative part generation results.}}
  \label{fig:part_gen}
\end{figure}

As illustrated in Fig.~\ref{fig:part_gen}, our method supports open-vocabulary,
text-grounded part generation across a diverse range of objects.
For each object, a natural-language query specifying the desired part is
sufficient to accurately extract the corresponding geometry.
The extracted parts exhibit high geometric fidelity to the input shape,
whether the target is a structurally complex object such as an octopus or a
geometrically simple one such as a wheel.
When multiple parts are extracted, they can be reassembled to yield a
compositional, part-by-part generation result.
Notably, although our approach delivers open-vocabulary text-grounded part
extraction, it is treated as a sub-task within a unified multimodal large
language model rather than as a dedicated pipeline, demonstrating the
generality of our framework.

\newpage

\section{Conclusion \& Future Work}

In this work, we present a unified multimodal 3D model that jointly addresses 3D understanding, 3D generation, and 3D editing within a single framework. Our model achieves state-of-the-art performance across all three tasks, and notably surpasses prior methods by a significant margin on both 3D generation and 3D editing.
We believe that unified 3D models represent a promising future direction, with the potential to address multiple 3D tasks within a single framework. However, several key challenges remain to be solved before this vision can be fully realized.

\begin{enumerate}
\item \textbf{Single-stage high-quality geometry representation.} The representation for 3D generation has not yet converged. To achieve high-quality geometry generation, current approaches often resort to multi-stage DiT pipelines---for example, TRELLIS~\cite{xiang2025structured} employs a multi-stage architecture. This multi-stage design poses significant challenges for unified models: if we wish to perform editing, we must also carry out multi-stage editing to obtain high-quality results, which is inherently difficult to scale up. Whether the 3D domain can achieve single-stage high-quality geometry generation---analogous to what has been accomplished in the image domain---remains an open and fundamental question.

\item \textbf{Text-to-3D \& 3D editing dataset captioning quality.} For text-to-3D generation, we currently rely on multimodal large language models such as Gemini for 3D asset captioning. However, these models still produce considerable ambiguity in their descriptions, leading to noisy training pairs in the text-to-3D dataset. We believe this issue will be progressively alleviated as multimodal large language models continue to advance.

\item \textbf{End-to-end texture editing.} Our current model primarily focuses on geometry editing. Texture editing remains largely unexplored due to the lack of suitable training data. This also raises a broader question: whether geometry and texture representations can be jointly modeled, and whether a unified single-stage representation for both high-quality geometry and texture can be achieved.

\item \textbf{Robust editing data construction pipeline.} Our current editing data construction pipeline, based on Nano3D-v2, maintains consistency by replacing tokens outside a box mask in the editing region. While this effectively preserves consistency in the unmasked exterior, the interior of the mask often contains non-edited regions whose consistency is difficult to guarantee. Such inconsistencies propagate into the end-to-end model during training, degrading editing quality. Developing a more robust and precise editing data construction pipeline remains an important open problem.

\item \textbf{New architecture exploration.} Our current framework adopts a cascaded AR + DiT architecture. In the image and video generation domains, Transfusion-style architectures have demonstrated remarkable effectiveness by deeply fusing information across multiple modalities. Exploring such architectures for 3D generation is a natural and promising next step.

\item \textbf{Data scaling.} We observe that both the quantity and quality of 3D data have not yet reached an ideal scale. Continuing to scale up the data---in both volume and quality---remains a critical direction for further improvement.
\end{enumerate}

\paragraph{Acknowledgement}
We thank the authors of \textbf{Omni123} for providing the results.
We thank \textbf{Zehuan Huang} for his contribution during his internship in Hunyuan.
We thank \textbf{Yifei Feng} and \textbf{Xin Yang} for sharing the texture generation codebase.

\newpage
\section{Author List}

Junliang Ye$^{*}$, Kenkun Liu$^{*}$, Guocun Wang$^{*}$, 
Yang Li$^{*\dagger}$, Yansong Qu$^{*}$, Chunshi Wang$^{*}$,\\
Jingwei Xu, Yunhan Yang, Zibo Zhao, Jiachen Xu, Jiaao Yu, Lifu Wang, Zhihao Liang, \\
Xin Huang, Zhuo Chen$^{\dagger}$, Chunchao Guo$^{\dagger}$ \\[0.9em]

{\small
\begin{tabular}{@{}r@{\hspace{0.7em}}l@{}}
$^{\dagger}$ & \textbf{Project Leaders:}
Yang Li, Zhuo Chen, and Chunchao Guo \\[0.35em]

$^{*}$ & \textbf{Core Contributors} \\[0.15em]

& \textit{3D Editing:}
Junliang Ye, Guocun Wang, Yansong Qu, Yang Li, Chunshi Wang, and Kenkun Liu \\[0.15em]

& \textit{Text-to-3D:}
Kenkun Liu, Junliang Ye, and Yang Li \\[0.15em]

& \textit{3D Understanding:}
Guocun Wang, Junliang Ye, Kenkun Liu, and Yang Li
\end{tabular}
}
\clearpage
\appendix

\section{Additional Results}

 \begin{figure}[h!]
  \centering
  \includegraphics[width=\linewidth]{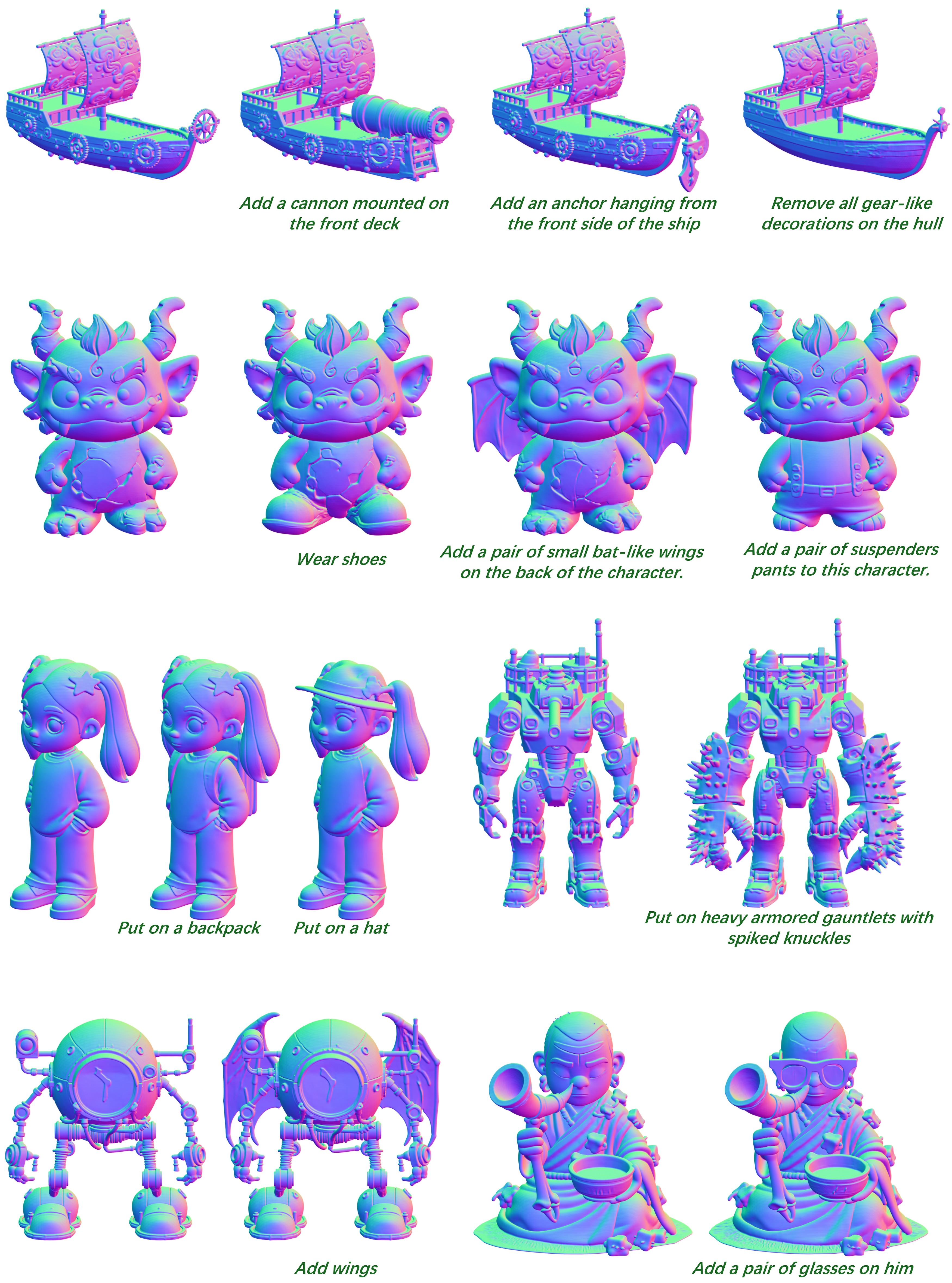}
  \caption{Qualitative shape editing results.}
  \label{fig:additional_edit_results}
\end{figure}

 \begin{figure}[h!]
  \centering
  \includegraphics[width=0.98\linewidth]{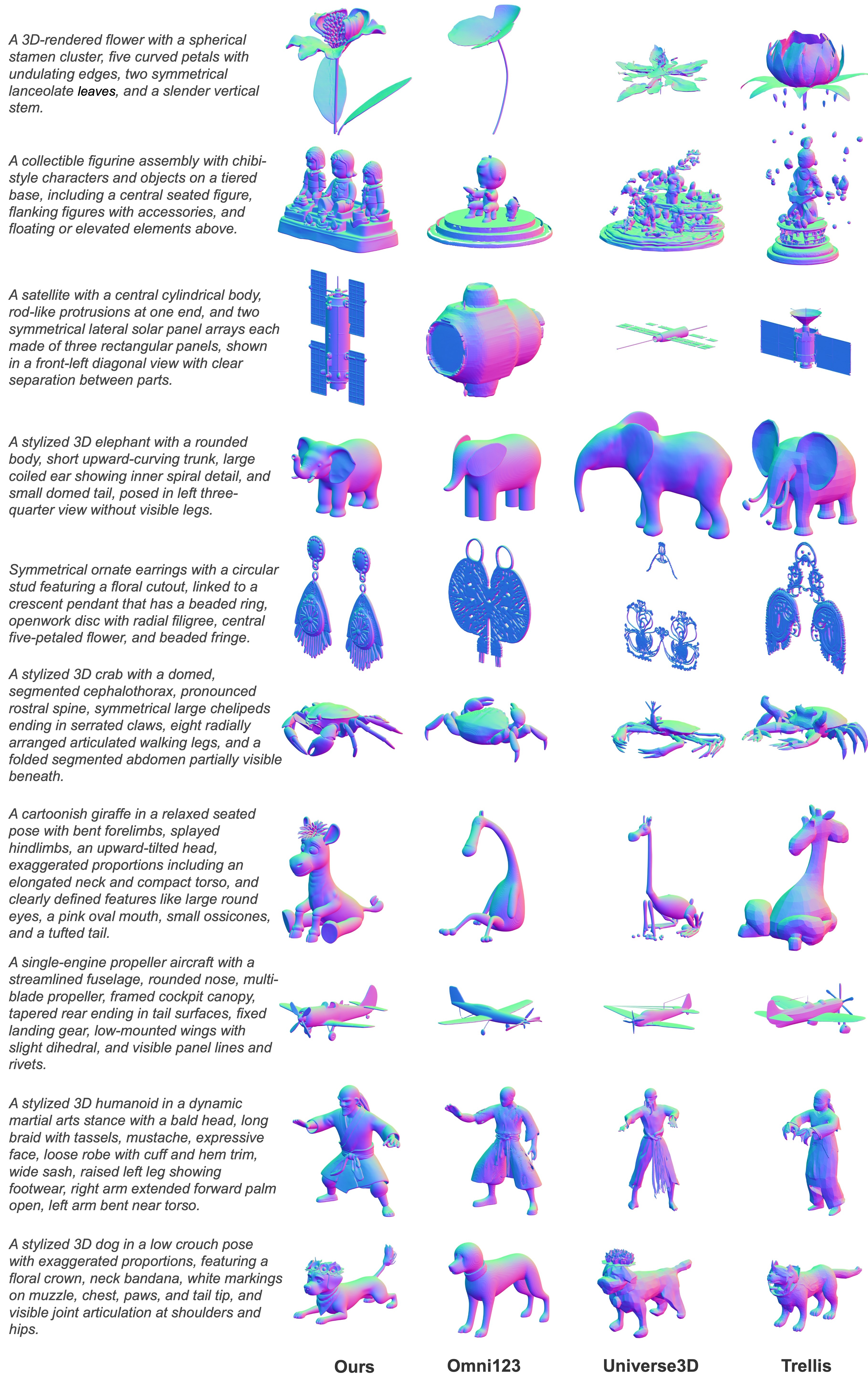}
  \caption{Qualitative text to 3D results.}
  \label{fig:t23d-compare2}
\end{figure}


\clearpage

\renewcommand{\refname}{References}
\renewcommand{\bibname}{References}
\renewcommand{\bibsection}{\section*{\raggedright \Large References}}
\makeatother

\bibliographystyle{abbrvnat}
\bibliography{egbib}

\end{document}